\documentclass[letterpaper]{article} 
\usepackage{aaai25}  
\usepackage{times}  
\usepackage{helvet}  
\usepackage{courier}  
\usepackage[hyphens]{url}  
\usepackage{graphicx} 
\usepackage{natbib}  
\usepackage{caption} 
\usepackage{algorithm}
\usepackage{algorithmic}
\usepackage{amssymb} 
\usepackage{amsmath}
\usepackage{multirow}
\usepackage{subfigure}
\usepackage{enumitem}
\usepackage{caption}
\usepackage{url}
\usepackage{array}
\usepackage{utfsym}
\newcommand{\mycomment}[1]{\textit{#1}}
\usepackage{newfloat}
\usepackage{listings}
\DeclareCaptionStyle{ruled}{labelfont=normalfont,labelsep=colon,strut=off} 
\floatstyle{ruled}
\newfloat{listing}{tb}{lst}{}
\floatname{listing}{Listing}
\title{Vision-aware Multimodal Prompt Tuning for Uploadable\\Multi-source Few-shot Domain Adaptation}
\author{
	Kuanghong Liu, Jin Wang\thanks{Corresponding author.}, Kangjian He\footnotemark[1], Dan Xu, Xuejie Zhang
}
\affiliations{
    School of Information Science and Engineering, Yunnan University, Kunming, China\\
    
    liukh@mail.ynu.edu.cn, \{wangjin, hekj, danxu, xjzhang\}@ynu.edu.cn
}

\usepackage{bibentry}

\begin{document}

\maketitle

\begin{abstract}
Conventional multi-source domain few-shot adaptation (MFDA) faces the challenge of further reducing the load on edge-side devices in low-resource scenarios. Considering the native language-supervised advantage of CLIP and the plug-and-play nature of prompt to transfer CLIP efficiently, this paper introduces an uploadable multi-source few-shot domain adaptation (UMFDA) schema. It belongs to a decentralized edge collaborative learning in the edge-side models that must maintain a low computational load. And only a limited amount of annotations in source domain data is provided, with most of the data being unannotated. Further, this paper proposes a vision-aware multimodal prompt tuning framework (VAMP) under the decentralized schema, where the vision-aware prompt guides the text domain-specific prompt to maintain semantic discriminability and perceive the domain information. The cross-modal semantic and domain distribution alignment losses optimize each edge-side model, while text classifier consistency and semantic diversity losses promote collaborative learning among edge-side models. Extensive experiments were conducted on OfficeHome and DomainNet datasets to demonstrate the effectiveness of the proposed VAMP in the UMFDA, which outperformed the previous prompt tuning methods. 
\end{abstract}
%
\begin{links}
    \link{Code}{https://github.com/lkh-meredith/VAMP-UMFDA}
\end{links}

\section{Introduction}
Multi-source few-shot domain adaptation (MFDA) is a resource-limited multi-source domain adaptation scenario, since large-scale manual annotations of each source domain are laborious and difficult, especially for the disease dataset that needs expert labeling or is even inaccessible when involves private data \cite{Gulshan2016,Harmon2020}. That means only a limited amount of annotations in the source domain data is provided, with most of the data being unannotated. Therefore, it is more practical and worth further exploration \cite{Yue2021a,Kim2020}. However, the improvement of the conventional MFDA method comes at the expense of storing additional clustering prototype features as a classifier \cite{Yue2021a}. In practice, this might increase the cost of the storage and computation for edge devices \cite{McMahan2016}.
\begin{figure}[!t]
    \centering
    \includegraphics[width=0.92\linewidth]{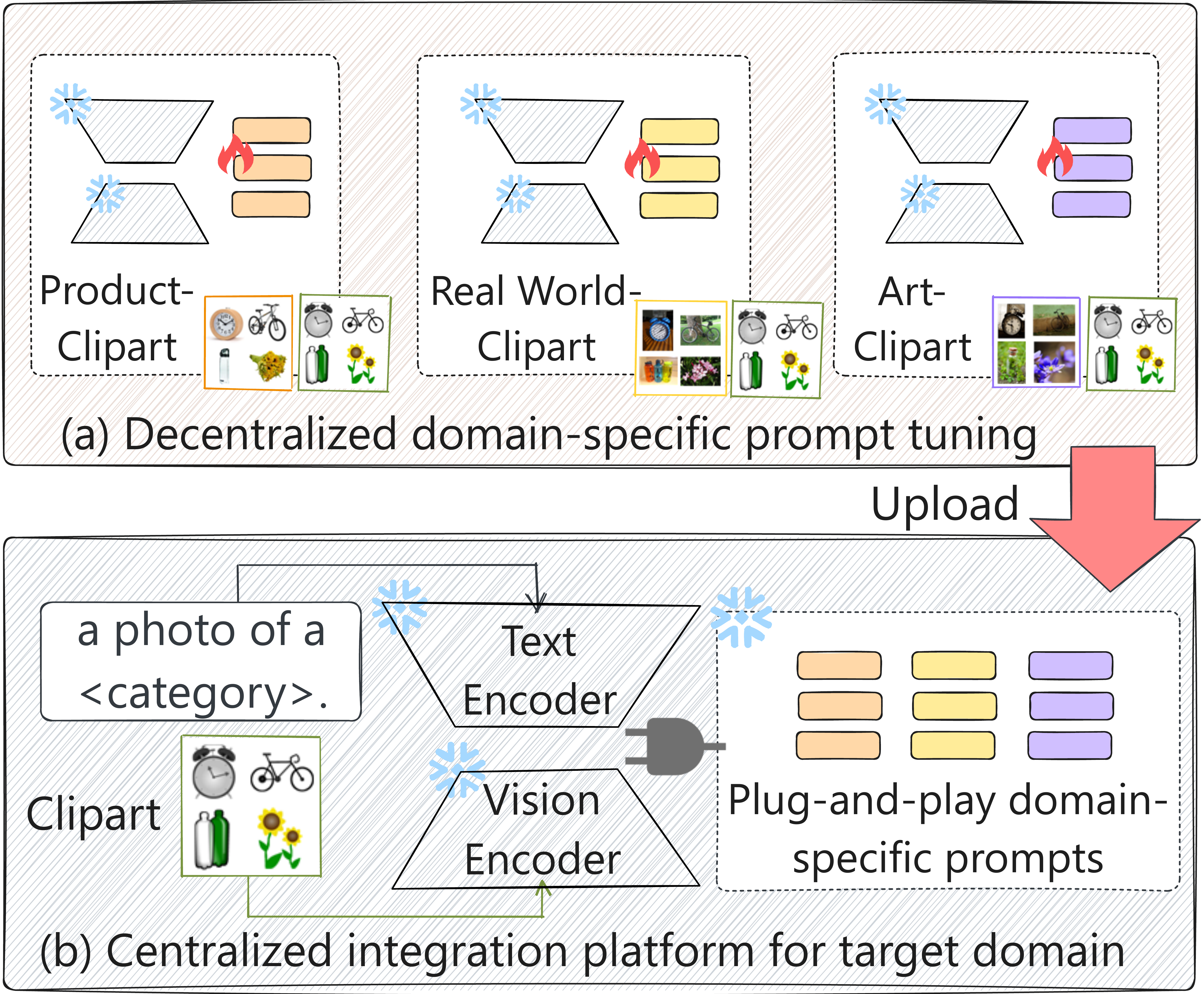}
    \caption{The illustration of uploadable multi-source few-shot domain adaptation (UMFDA) schema for decentralized edge learning.}
    \label{fig:fig1}
\end{figure}

The pretrained vision-and-language model (VLM), such as CLIP \cite{Radford2021}, has attracted attention for its remarkable zero-shot inference performance and transferability. Pretraining by cross-modal alignment of contrastive learning, CLIP has the native advantage of language-guided supervision to form similar semantic clusters. A critical insight is to leverage a manual-craft text prompt, e.g., \texttt{a photo of a <category>.}, as a query prompt for the text encoder to inspire CLIP's potential. With the development of prompt tuning in NLP \cite{Lester2021,Li2021,Liu2021,Shin2020}, some methods \cite{Zhou2022b,Zang2022,Chen2022,Lu2022,Khattak2023} replaced the manual-crafted prompt with a small fraction of learnable parameters. Only by updating the fewer prompt parameters rather than the entire model will these plug-and-play learnable prompts make CLIP favorably adaptable to various downstream tasks, even under the few annotated sample conditions. With insights into the native language-supervised benefit of CLIP and the prompt's efficient plug-and-play capability, they are well-suited for application in low-resource, low-load edge computing scenarios. Therefore, we introduce an uploadable multi-source few-shot domain adaptation (UMFDA) schema for decentralized edge learning in this paper, as illustrated in Figure \ref{fig:fig1}.
\begin{figure*}[t]
    \centering
    \includegraphics[width=0.98\linewidth]{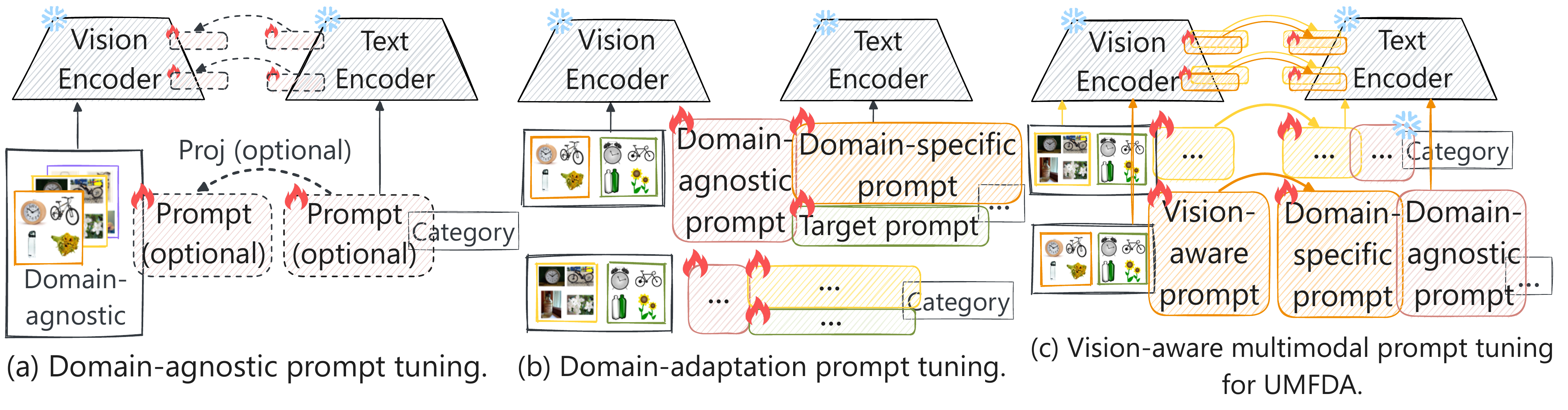}
    \caption{Summary of various prompt tuning technologies (best viewed in color). (a) concludes the several prevalent prompt tuning methods while they are domain-agnostic. (b) represents the typical prompt tuning methods of single-source domain adaptation focusing on disentangling the prompts to explore the difference between the source and target domains. It must be further aligned in the center device among multiple source domains. (c) is our proposed vision-aware multimodal prompt tuning method tailored for the UMFDA.}
    \label{fig:fig2}
\end{figure*}

Specifically, Figure \ref{fig:fig1}(a) is the decentralized training stage for edge-side devices. By domain-specific prompt tuning, these edge-side devices perform lightweight edge collaborative learning. It avoids the need to finetune the entire domain-specific model and reduces the burden of storing and processing trainable parameters \cite{Zhao2024}. Figure \ref{fig:fig1}(b) is the centralized integration platform that accepts all the uploaded domain-specific prompts. Once these trained domain-specific prompts are inserted into the frozen CLIP, the domain-specific image extractor and text classifier are constructed, where the image encoder of CLIP serves as a feature extractor; the text encoder of CLIP is regarded as a text classifier. The target domain's results are integrated from directly inferences of all the domain-specific models without further training in the centralized platform.

Nevertheless, the current prompt tuning technologies are not suitable for the UMFDA. As summarized in Figure \ref{fig:fig2}(a), the prevalent prompt tuning methods \cite{Zhou2022b,Jia2022,Khattak2023} are domain-agnostic and neglect domain shifts and distribution differences among domains. Figure \ref{fig:fig2}(b) represents another recently emerged domain adaption prompt tuning method \cite{Ge2022,Chen2023}. They disentangle context prompts as domain-agnostic and domain-specific prompts to embed domain information into text prompts by contrasting the source and target domain data pair. However, this approach increases the central equipment's computing load, as additional training is still required to centrally derive a domain-invariant shared representation space among the learned individual prompts \cite{Chen2023}. Additionally, they only affect the text encoder when adapting to changing domains, while learning domain-specific discriminant image features is difficult. It may damage the distribution of the representation in CLIP and cause a loss of the learned semantic information \cite{Singha2023,Bai2024,Du2024}.

This study proposes a vision-aware multimodal prompt tuning framework (VAMP) for the UMFDA, as shown in Figure \ref{fig:fig2}(c). The multimodal prompt tuning is applied to maintain the discriminative ability of the learned features. Notably, the difference from the previous prompt-agnostic multimodal prompt (\textit{text → vision}, Figure \ref{fig:fig2}(a)) is that the visual prompt perceives the domain information and is then projected to the text prompt used for the specific domain (\textit{vision → text}). By doing so, these domain-specific text prompts guided by visual perception can be searched to describe the domain images, rather than being the manual designing or using vision-independent text prompts. The original manual text prompt, i.e., \texttt{A photo of a <category>.}, as the domain-agnostic prompt is concatenated with the domain-specific text prompt to preserve the generalization knowledge of CLIP.

To optimize this framework in a decentralized manner, the pairs source and target domain data are fully utilized. Each edge-side model with domain-specific vision-aware multimodal prompts is trained internally, while engaging in collaborative learning among the edge-side models. Concretely, cross-modal semantic alignment (CSA) and domain distribution alignment (DDA) losses are used in each edge-side model; text classifier consistency (TCC) and text semantic diversity (TSD) losses are introduced to facilitate collaborative learning among multiple edge-side models. The main contribution can be summarized as follows:
\begin{itemize}
\item Inspired by the plug-and-play prompts to transfer the CLIP efficiently, this study introduces a UMFDA schema for low-resource and low-load edge learning and further proposes the VAMP framework. The customized domain-specific prompts in the VAMP are vision-aware multimodal prompts, where vision-aware prompts guide domain-specific text prompts.
\item VAMP is optimized in a decentralized training manner by four different losses. CSA and DDA losses ensure cross-modal semantic information and distribution alignments within edge-side models; TCC and TSD losses facilitate collaborative learning among edge-side models.
\item Extensive experiments on OfficeHome and DomainNet datasets demonstrate the effectiveness of the VAMP, which outperforms the previous prompt tuning methods.
\end{itemize} 
\section{Preliminaries}
\noindent \textbf{MFDA.} The UMFDA also follows the setting of MFDA from MFSAN \cite{Yue2021a}, which focuses on transferring generalization knowledge from the multiple source domains to the target domain. In the scenario, there is a small annotated data $D_{s\_a}^i = \{ (x_j^{a,i},y_j^{a,i})\} _{j = 1}^{N_a^i}$ and a large unannotated data $D_{s\_u}^i = \{ x_j^{u,i}\} _{j = 1}^{N_u^i}$ for the $i$-th source domain $D_s^i = D_{s\_a}^i \cup D_{s\_u}^i,i \in \{ 1,2,...,M\}$, where ${N_a^i}$ and $N_u^i$ ($N_a^i \ll N_u^i$) are the size of the annotated and unannotated samples, respectively. $M$ is the number of source domains. ${D_t} = \{ x_j^t\} _{j = 1}^{{N_t}}$ denotes the target dataset without label annotation, where ${N_t}$ represents the count of target samples. Notably, it assumes that the data of different domains come from different distributions and all share the same label space. The objective is to train a domain adaptation model on multiple domain data $D_s$ and $D_t$, which enables prediction labels on the target samples as correctly as possible.

\noindent \textbf{CLIP Inference.}
Benefiting from the alignment pretraining of language and vision modalities on large-scale text-image pairs by contrastive learning, the CLIP model has achieved outstanding zero-shot inference. A piece of manual-crafted text prompts, i.e., \texttt{a photo of a <category>.}, is enough to inspire CLIP's potential \cite{Radford2021}. Given an image $x$, its visual representation $z \in \mathbb{R}^d$ and text representations $W = \{ {w_1},{w_2},...,{w_K}\}  \in {\mathbb{R}^{K \times d}}$ for the $K$ candidate categories are produced by the CLIP’s vision encoder $\Psi$ and text encoder $\Phi$, respectively. The probability that the image belongs to the $c$-th class is calculated as, 
\begin{equation}
    p(\hat y = c|x) = \frac{{\exp (\cos ({w_c},z)/{\rm T} )}}{{\sum\limits_{k = 1}^K {\exp (\cos ({w_k},z)/{\rm T})} }}
    \label{eq:eq1}
\end{equation}
where $\cos ( \cdot , \cdot )$ denotes the cosine similarity and $\rm T$ is a fixed temperature parameter learned by CLIP.

\noindent \textbf{Multimodal Prompt Tuning.}
MaPLe \cite{Khattak2023} introduced a coupling function to enhance interaction with the prompts of vision and text modality for synergic optimization in the CLIP. Specifically, a series of new learnable prompts ${\bf{p}} = [{{\bf{p}}_1},{{\bf{p}}_2},...,{{\bf{p}}_b}] \in {\mathbb{R}^{b \times {d_T}}}$ are introduced in each transformer layer of $\Phi$, up to the depth $J$. Vision prompts ${{\bf{\tilde p}}^l}$ are obtained by projecting ${{\bf{p}}^l}$ via a coupling function ${\rm{pro}}{{\rm{j}}^l}( \cdot )$ at $l$-th transformer layer, i.e., ${{\bf{\tilde p}}^l}{\rm{ = pro}}{{\rm{j}}^l}{\rm{(}}{{\bf{p}}^l})$. The ${\rm{pro}}{{\rm{j}}^l}( \cdot )$ is implemented as one linear layer that maps dimension $d_T$ to $d_V$. Supposing that both encoder $\Phi$ and $\Psi$ have $L$ transformer layers and image patch embeddings are ${\bf{q}^0} = [{{\bf{q}}_1^0},{{\bf{q}}_2^0},...,{{\bf{q}}_s^0}] \in {\mathbb{R}^{s \times {d_V}}}$, the entire process can be written as,
\begin{equation}
\small
\begin{aligned}
[{\bf{c}}_T^l, \_, {{\bf{e}}^l}] &= f_\Phi^l([{\bf{c}}_T^{l - 1}, {{\bf{p}}^{l - 1}}, {{\bf{e}}^{l - 1}}]), \; l = 1, 2, \ldots, J \\
[{\bf{c}}_T^m, {{\bf{p}}^m}, {{\bf{e}}^m}] &= f_\Phi^m([{\bf{c}}_T^{m - 1}, {{\bf{p}}^{m - 1}}, {{\bf{e}}^{m - 1}}]), \; m = J + 1, \ldots, L \\
w &= {\rm{Pro}}{{\rm{j}}_\Phi}({e}_n^L)
\end{aligned}
\label{eq:eq2}
\end{equation}
\begin{equation}
\small
\begin{aligned}
[{\bf{c}}_V^l,{{\bf{q}}^l},\_] &= f_\Psi ^l([{\bf{c}}_V^{l - 1},{{\bf{q}}^{l - 1}},{\rm{pro}}{{\rm{j}}^{l - 1}}{\rm{(}}{{\bf{p}}^{l - 1}})]),l = 1,2,...,J\\
[{\bf{c}}_V^m,{{\bf{q}}^m},{{{\bf{\tilde p}}}^m}] &= f_\Psi ^m([{\bf{c}}_V^{m - 1},{{\bf{q}}^{m - 1}},{{{\bf{\tilde p}}}^{m - 1}}]),m = J + 1,...,L\\
z &= {\rm{Pro}}{{\rm{j}}_\Psi }({\bf{c}}_V^L)
\end{aligned}
\label{eq:eq3}
\end{equation}
where ${{\bf{c}}_T} \in {\mathbb{R}^{{d_T}}}$ and ${{\bf{c}}_V} \in {\mathbb{R}^{{d_V}}}$ are the starting token embeddings. ${\bf{e}^0} = [{{e}_1^0},{{e}_2^0},...,{{e}_n^0}] \in {\mathbb{R}^{n \times {d_T}}}$  are the fixed token embeddings including category label. $[ \cdot , \cdot ]$ represents the concatenation operation. Each text representation $w \in {\mathbb{R}^d}$ and image representation $z \in {\mathbb{R}^d}$ are finally projected to a common vision-and-language space via ${\rm{Pro}}{{\rm{j}}_\Phi }( \cdot )$ and ${\rm{Pro}}{{\rm{j}}_\Psi }( \cdot )$, respectively.
\section{Method}
\subsection{Vision-aware Multimodal Prompt Tuning}
In the UMFDA setting, we still adopt multimodal prompts as domain-specific prompts to maintain semantic discriminability. Differently, the domain information is perceived by visual prompts during tuning the image extractor. Then, these visual prompts are projected as learnable domain-specific text prompts concatenated with the manual-crafted prompt to search optimal text queries in the corresponding domain. Therefore, the reverse of the coupling function in MaPLe, i.e.,${{\bf{p}}^l}{\rm{ = pro}}{{\rm{j}}^l}{\rm{(}}{{\bf{\tilde p}}^l})$, our vision-aware prompt tuning is written as,
\begin{equation}
\small
\begin{aligned}
[{\bf{c}}_V^l,{{\bf{q}}^l},\_] &= f_\Psi ^l([{\bf{c}}_V^{l - 1},{{\bf{q}}^{l - 1}},{{{\bf{\tilde p}}}^{l - 1}}]),l = 1,2,...,J\\
[{\bf{c}}_V^m,{{\bf{q}}^m},{{{\bf{\tilde p}}}^m}] &= f_\Psi ^m([{\bf{c}}_V^{m - 1},{{\bf{q}}^{m - 1}},{{{\bf{\tilde p}}}^{m - 1}}]),m = J + 1,...,L\\
z &= {\rm{Pro}}{{\rm{j}}_\Psi }({\bf{c}}_V^L)
\end{aligned}
\label{eq:eq4}
\end{equation}
\begin{equation}
\small
\begin{aligned}
[{\bf{c}}_T^l,\_,{{\bf{e}}^l}] &= f_\Phi ^l([{\bf{c}}_T^{l - 1},{\rm{pro}}{{\rm{j}}^{l - 1}}{\rm{(}}{{{\bf{\tilde p}}}^{l - 1}}),{{\bf{e}}^{l - 1}}]),l = 1,2,...,J\\
[{\bf{c}}_T^m,{{\bf{p}}^m},{{\bf{e}}^m}] &= f_\Phi ^m([{\bf{c}}_T^{m - 1},{{\bf{p}}^{m - 1}},{{\bf{e}}^{m - 1}}]),m = J + 1,...,L\\
w &= {\rm{Pro}}{{\rm{j}}_\Phi }({e}_n^L)
\end{aligned}
\label{eq:eq5}
\end{equation}
where ${{\bf{e}}^0}$ denotes the fixed embeddings of the original text prompt, i.e., ``\texttt{a photo of a <category>.}'', which is seen as the domain-agnostic prompt.
\begin{figure*}[t]
    \centering
    \includegraphics[width=0.97\linewidth]{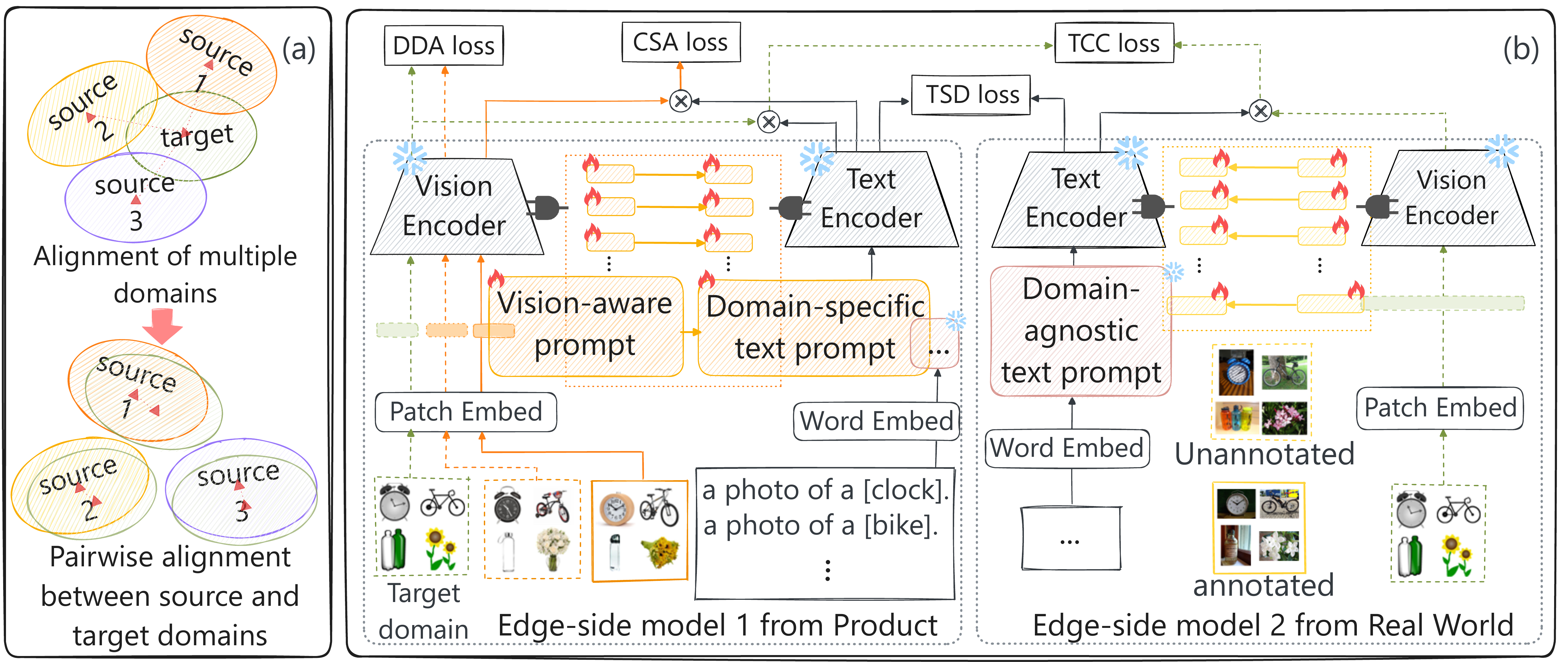}
    \caption{(a) Illustration of a conceptual diagram of domain alignment. The first alignment approach at the top, multiple source domains aligning with the target domain, is unsuitable for the decentralized edge computing scenario because the centralized model needs more aligning training. It is also hard to match all distributions of source and target domains. The second idea inspired by Zhu et al. \shortcite{Zhu2019}, pair-wise alignment between source and target domains, is thus adopted by the VAMP framework. (b) The proposed decentralized training framework VAMP. For clarity, only two edge-side models are drawn here.}
    \label{fig:fig3}
\end{figure*}
\subsection{VAMP Framework}
Figure \ref{fig:fig3} is the conceptual diagram of the proposed VAMP framework, which aims to learn a group of trainable domain-specific vision-aware multimodal prompts $\{ {{\bf{P}}_1},{{\bf{P}}_2},...,{{\bf{P}}_M}\}$ (including newly inserted learnable prompts for each layer and their corresponding projection functions) for different edge-side models in decentralized training way. Domain-specific image extractor and text classifier for source domain $i$ are thus represented as ${\Psi _i}( \cdot ,{{\bf{P}}_i})$ and ${\Phi _i}({\bf{E}}, {{\bf{P}}_i})$, respectively, where ${\bf{E}} = \{ {{\bf{e}}_1},{{\bf{e}}_2},...,{{\bf{e}}_K}\} \in {\mathbb{R}^{K \times n \times {d_T}}}$ denotes the fixed token embeddings of all categories. In this framework, $M$ domain-alignment vision representation spaces are learned by $\{ {\Psi _i}( \cdot ,{{\bf{P}}_i})\} _{{i = 1}}^M$, rather than learning a common domain-invariant feature space by aligning multiple domains in the subsequent centralized integration stage, as illustrated in Figure \ref{fig:fig3} (a). The overall decentralized training framework VAMP is shown in Figure \ref{fig:fig3} (b), where the CSA and DDA losses are used in each edge-side model; the TCC and TSD losses are introduced to facilitate collaborative learning among multiple edge-side models. They are described in detail as follows.

\noindent \textbf{Cross-modal Semantic Alignment.} For each domain-specific model, a batch of annotated data $\{ (x_j^{a,i},y_j^{a,i})\} _{j = 1}^{n_a^i} \subset D_{s\_a}^i$ and unannotated target data $\{ x_j^t\} _{j = 1}^{{n_t}} \subset {D_t}$ are sampled and applied to keep the semantic discriminability by cross-modal contrastive learning. For the annotated source samples, the model can be directly optimized with ground-truth labels as, 
\begin{equation}
\small
{\cal L}_{CSA}^a =  - \frac{1}{{n_a^i}}\sum\limits_{j = 1}^{n_a^i} {\sum\limits_{k = 1}^K {\left[ {y_{j,k}^{a,i} \cdot \log p(\hat y_{j,k}^{a,i}|x_j^{a,i};{{\bf{P}}_i})} \right]} }
\label{eq:eq6}
\end{equation}
where $y_{j}^{a,i}$ is the one-hot ground-truth label. $p(\hat y = c|x;{{\bf{P}}_i})$ of an image $x$ categorizing to the $c$-th class is rewritten as,
\begin{equation}
\small
p(\hat y = c|x;{{\bf{P}}_i}) = \frac{{\exp (\cos (\Phi ({{\bf{E}}_c},{{\bf{P}}_i}),\Psi (x,{{\bf{P}}_i}))/{\rm T})}}{{\sum\limits_{k = 1}^K {\exp (\cos (\Phi ({{\bf{E}}_k},{{\bf{P}}_i}),\Psi (x,{{\bf{P}}_i}))/{\rm T})} }}
\label{eq:eq7}
\end{equation}
Meanwhile, by zero-shot inference of CLIP as shown in Eq.(\ref{eq:eq1}), a qualified pseudo label of target sample whose maximum prediction probability $\hat\tau$ is larger than a fixed threshold ${\cal T}$ is generated. When the number of qualified pseudo labels is not equal to 0 in a batch,
\begin{equation}
    \small
    {\cal L}_{CSA}^t =  - \frac{{\sum\limits_{j = 1}^{{n_t}} {\mathbb{I} \{ {{\hat \tau }_j} \ge {\cal T}\} \sum\limits_{k = 1}^K {\left[ {\hat y_{j,k}^{{\rm zs}} \cdot \log {p_i}(\hat y_{j,k}^t|x_j^t;{{\bf{P}}_i})} \right]} } }}{{\sum\limits_{j = 1}^{{n_t}} {\mathbb{I}\{ {{\hat \tau }_j} \ge {\cal T}\} } }}
    \label{eq:eq8}
\end{equation}
where $\mathbb{I}\{ \cdot \} $ is an indicator function. $\hat y_{j}^{\rm zs}$ is a one-hot pseudo label predicted by zero-shot CLIP. Therefore, the cross-modal semantic alignment loss is written as,
\begin{equation}
 \small
{{\cal L}_{CSA}} = {\cal L}_{CSA}^a + {\cal L}_{CSA}^t
\label{eq:eq9}
\end{equation}

\noindent \textbf{Domain Distribution Alignment.} To align the distribution for each pair of source and target domain, maximum mean discrepancy (MMD) \cite{Gretton2007,Gretton2012} is used to optimize by exploiting extensive unannotated data from $D_{s\_u}^i$ and ${D_t}$. Supposed that the image features $\{ z_j^{u,i}\} _{j = 1}^{N_u^i} = \{ {\Psi _i}(x_j^{u,i},{{\bf{P}}_i})\} _{j = 1}^{N_u^i}$ and $\{ z_j^{t}\} _{j = 1}^{{N_t}} = \{ {\Psi _i}(x_j^t,{{\bf{P}}_i})\} _{j = 1}^{{N_t}}$ are independently and identically drawn from $P_s^i$ and ${Q_t}$, respectively. MMD works by distinguishing statistical hypothesis testing of two samples that if they are similar then they are likely from the same distribution. To measure the difference between $P_s^i$ and ${Q_t}$, the squared MMD with kernel mean embeddings is formulated as,
\begin{equation}
\small
{\rm{MM}}{{\rm{D}}^2}({\cal F},P_s^i,{Q_t}) \!\buildrel \Delta \over =\! \left\| {{\mathbb{E}_{{z^{u,i}}\sim P_s^i}}\left[ {\phi ({z^{u,i}})} \right] \!-\! {\mathbb{E}_{{z^t}\sim {Q_t}}}\left[ {\phi ({z^t})} \right]} \right\|_{\cal H}^2
\label{eq:eq10}
\end{equation}
where the function class $\cal F$ is the unit ball in a  reproducing kernel Hilbert space (RKHS) $\cal H$ endowed with a characteristic kernel $\kappa$. $\phi ( \cdot )$ is feature mapping that maps into $\cal H$ and $\kappa $ denotes $\kappa ({z^{u,i}},{z^t}) = {\left\langle {\phi ({z^{u,i}}),\phi ({z^t})} \right\rangle _{\cal H}}$, where $\left\langle { \cdot , \cdot } \right\rangle $ is the inner product of vectors. Practically, given a batch $n_u^i$ of source unannotated samples and ${n_t}$ target samples, the form of empirical estimates with finite samples is calculated as,
\begin{equation}
\small
\begin{aligned}
{\widehat {{\rm{MMD}}}^2}({\cal F},P_s^i,{Q_t}) = \frac{1}{{n_u^i \cdot n_u^i}}\sum\limits_{j = 1}^{n_u^i} {\sum\limits_{k \ne j}^{n_u^i} {\kappa (z_j^{u,i},z_k^{u,i})} } \\
 + \frac{1}{{{n_t} \cdot {n_t}}}\sum\limits_{j = 1}^{{n_t}} {\sum\limits_{k \ne j}^{{n_t}} {\kappa (z_j^t,z_k^t)} }  - \frac{2}{{n_u^i \cdot {n_t}}}\sum\limits_{j = 1}^{n_u^i} {\sum\limits_{k = 1}^{{n_t}} {\kappa (z_j^{u,i},z_k^t)} } 
\end{aligned}
\label{eq:eq11}
\end{equation}
where $\kappa$ is the universal kernel such as Gaussian kernel $\kappa (z,z') = \exp \left( { - \frac{1}{{2\sigma }}{{\left| {z - z'} \right|}^2}} \right)$ with bandwidth $\sigma $. When the feature space is a universal RKHS, MMD is 0 if and only if $P_s^i = {Q_t}$. Therefore, minimizing MMD under this condition is equivalent to minimizing the distance between all moments of the two distributions $P_s^i$ and ${Q_t}$ \cite{Li2015,Guo2020}. To achieve domain distribution alignment, MMD is expected to become smaller and smaller between the sampled feature distribution from the pair of source and target domains by tuning vision-aware prompts. Here, the squared MMD as defined in Eq. (\ref{eq:eq11}) is applied as,
 \begin{equation}
 \small
{{\cal L}_{DDA}} = {\widehat {{\rm{MMD}}}^2}({\cal F},P_s^i,{Q_t})
\label{eq:eq12}
 \end{equation}
\textbf{Text Classifier Consistency.} Different domain-specific text classifiers' predictions on the edge sides are expected to be consistent when inputting the same unobserved target samples. The decentralized consistency and collaboration among these edge-side models are realized by doing so. The discrepancy among all domain-specific text classifiers is minimized by,
\begin{equation}
\resizebox{\linewidth}{!}{
${{\cal L}_{TCC}} = \frac{1}{{{n_t} K}{{\cal C}_M^2}}\sum\limits_{i = 1}^{M - 1} {\sum\limits_{r = i + 1}^M {\sum\limits_{j = 1}^{{n_t}} {\sum\limits_{k = 1}^K {\left| {{p_i}(\hat y_{j,k}^t|x_j^t;{{\bf{P}}_i}) - {p_r}(\hat y_{j,k}^t|x_j^t;{{\bf{P}}_r})} \right|} } } }$
}
\label{eq:eq13}
\end{equation}
where the combination ${\cal C}_M^2$ represents the number of distinct pairs that can be formed from $M$ source domains.
\begin{table*}[!t]
\centering
\small
\setlength{\tabcolsep}{1.2mm} 
\renewcommand{\arraystretch}{1.0}
\begin{tabular}{cccccccccccc}
\hline
\multicolumn{12}{c}{OfficeHome} \\ \hline
\multicolumn{2}{c|}{\multirow{3}{*}{Method}} &
  \multicolumn{5}{c|}{3\%} &
  \multicolumn{5}{c}{6\%} \\ \cline{3-12} 
\multicolumn{2}{c|}{} &
  \multirow{2}{*}{\begin{tabular}[c]{@{}c@{}}APR\\ →C\end{tabular}} &
  \multirow{2}{*}{\begin{tabular}[c]{@{}c@{}}CPR\\ →A\end{tabular}} &
  \multirow{2}{*}{\begin{tabular}[c]{@{}c@{}}CAR\\ →P\end{tabular}} &
  \multirow{2}{*}{\begin{tabular}[c]{@{}c@{}}CAP\\ →R\end{tabular}} &
  \multicolumn{1}{c|}{\multirow{2}{*}{Avg}} &
  \multirow{2}{*}{\begin{tabular}[c]{@{}c@{}}APR\\ →C\end{tabular}} &
  \multirow{2}{*}{\begin{tabular}[c]{@{}c@{}}CPR\\ →A\end{tabular}} &
  \multirow{2}{*}{\begin{tabular}[c]{@{}c@{}}CAR\\ →P\end{tabular}} &
  \multirow{2}{*}{\begin{tabular}[c]{@{}c@{}}CAP\\ →R\end{tabular}} &
  \multirow{2}{*}{Avg} \\
\multicolumn{2}{c|}{} &
   &
   &
   &
   &
  \multicolumn{1}{c|}{} &
   &
   &
   &
   &
   \\ \hline
\multicolumn{1}{c|}{Zero-shot} &
  \multicolumn{1}{c|}{CLIP} &
  67.7 &
  82.7 &
  89.2 &
  89.6 &
  \multicolumn{1}{c|}{82.3} &
  67.7 &
  82.7 &
  89.2 &
  89.6 &
  82.3 \\ \hline
\multicolumn{1}{c|}{\multirow{3}{*}{\begin{tabular}[c]{@{}c@{}}Domain-agnostic\\ prompt\end{tabular}}} &
  \multicolumn{1}{c|}{CoOP} &
  70.0±0.3 &
  82.4±0.6 &
  90.9±0.2 &
  90.3±0.3 &
  \multicolumn{1}{c|}{83.4±0.2} &
  70.3±0.6 &
  82.5±0.4 &
  90.6±0.4 &
  90.4±0.8 &
  83.5±0.5 \\
\multicolumn{1}{c|}{} &
  \multicolumn{1}{c|}{VPT} &
  69.6±0.4 &
  83.0±0.5 &
  90.2±0.2 &
  90.2±0.1 &
  \multicolumn{1}{c|}{83.3±0.2} &
  69.7±0.4 &
  83.9±0.6 &
  90.5±0.4 &
  90.3±0.4 &
  83.6±0.3 \\
\multicolumn{1}{c|}{} &
  \multicolumn{1}{c|}{MaPLe} &
  71.1±0.7 &
  83.3±0.5 &
  91.2±0.2 &
  90.5±0.6 &
  \multicolumn{1}{c|}{84.0±0.4} &
  71.0±0.6 &
  83.1±0.5 &
 \textbf{91.8}±0.4 &
  90.7±0.2 &
  84.1±0.3 \\ \hline
\multicolumn{1}{c|}{\multirow{2}{*}{\begin{tabular}[c]{@{}c@{}}Disentangling\\ prompt\end{tabular}}} &
  \multicolumn{1}{c|}{DAPL} &
  70.0±0.1 &
  83.5±0.7 &
  91.0±0.5 &
  90.5±0.2 &
  \multicolumn{1}{c|}{83.7±0.4} &
  70.2±0.3 &
  84.3±0.3 &
  90.9±0.4 &
  90.6±0.2 &
  84.0±0.1 \\
\multicolumn{1}{c|}{} &
  \multicolumn{1}{c|}{MPA} &
  63.0±0.5 &
  76.9±1.3 &
  83.5±1.1 &
  81.6±0.4 &
  \multicolumn{1}{c|}{76.2±0.2} &
  63.5±0.7 &
  77.3±0.9 &
  83.4±0.3 &
  81.2±0.5 &
  76.3±0.3 \\ \hline
\multicolumn{2}{c|}{VAMP} &
  \textbf{73.7}±0.6 &
  \textbf{85.7}±0.3 &
  \textbf{91.4}±0.4 &
  \textbf{90.9}±0.2 &
  \multicolumn{1}{c|}{\textbf{85.4}±0.2} &
  \textbf{73.5}±0.2 &
  \textbf{85.8}±0.4 &
  91.4±0.1 &
  \textbf{91.4}±0.2 &
  \textbf{85.5}±0.1 \\ \hline
\multicolumn{12}{c}{DomainNet} \\ \hline
\multicolumn{2}{c|}{\multirow{3}{*}{Method}} &
  \multicolumn{5}{c|}{1 shot} &
  \multicolumn{5}{c}{3 shot} \\ \cline{3-12} 
\multicolumn{2}{c|}{} &
  \multirow{2}{*}{\begin{tabular}[c]{@{}c@{}}PRS\\ →C\end{tabular}} &
  \multirow{2}{*}{\begin{tabular}[c]{@{}c@{}}CRS\\ →P\end{tabular}} &
  \multirow{2}{*}{\begin{tabular}[c]{@{}c@{}}CPS\\ →R\end{tabular}} &
  \multirow{2}{*}{\begin{tabular}[c]{@{}c@{}}CPR\\ →S\end{tabular}} &
  \multicolumn{1}{c|}{\multirow{2}{*}{Avg}} &
  \multirow{2}{*}{\begin{tabular}[c]{@{}c@{}}PRS\\ →C\end{tabular}} &
  \multirow{2}{*}{\begin{tabular}[c]{@{}c@{}}CRS\\ →P\end{tabular}} &
  \multirow{2}{*}{\begin{tabular}[c]{@{}c@{}}CPS\\ →R\end{tabular}} &
  \multirow{2}{*}{\begin{tabular}[c]{@{}c@{}}CPR\\ →S\end{tabular}} &
  \multirow{2}{*}{Avg} \\
\multicolumn{2}{c|}{} &
   &
   &
   &
   &
  \multicolumn{1}{c|}{} &
   &
   &
   &
   &
   \\ \hline
\multicolumn{1}{c|}{Zero-shot} &
  \multicolumn{1}{c|}{CLIP} &
  82.7 &
  82.6 &
  91.8 &
  79.6 &
  \multicolumn{1}{c|}{84.2} &
  82.7 &
  82.6 &
  91.8 &
  79.6 &
  84.2 \\ \hline
\multicolumn{1}{c|}{\multirow{3}{*}{\begin{tabular}[c]{@{}c@{}}Domain-agnostic\\ prompt\end{tabular}}} &
  \multicolumn{1}{c|}{CoOP} &
  82.3±0.6 &
  81.9±0.5 &
  90.9±0.4 &
  79.1±0.3 &
  \multicolumn{1}{c|}{83.5±0.4} &
  83.5±0.2 &
  82.8±0.3 &
  91.1±0.1 &
  80.0±0.6 &
  84.3±0.3 \\
\multicolumn{1}{c|}{} &
  \multicolumn{1}{c|}{VPT} &
  82.4±0.2 &
  82.2±0.3 &
  91.7±0.0 &
  80.0±0.2 &
  \multicolumn{1}{c|}{84.0±0.1} &
  83.4±0.1 &
  83.1±0.1 &
  91.9±0.1 &
  80.3±0.0 &
  84.7±0.0 \\
\multicolumn{1}{c|}{} &
  \multicolumn{1}{c|}{MaPLe} &
  83.3±0.6 &
  82.9±0.0 &
  91.8±0.3 &
  79.6±0.2 &
  \multicolumn{1}{c|}{84.4±0.1} &
  84.6±0.3 &
  83.3±0.2 &
  91.6±0.2 &
  80.1±0.5 &
  84.9±0.1 \\ \hline
\multicolumn{1}{c|}{\multirow{2}{*}{\begin{tabular}[c]{@{}c@{}}Disentangling\\ prompt\end{tabular}}} &
  \multicolumn{1}{c|}{DAPL} &
  83.4±0.8 &
  83.7±0.3 &
  91.9±0.4 &
  80.8±0.1 &
  \multicolumn{1}{c|}{85.0±0.4} &
  83.6±0.4 &
  84.3±0.2 &
  92.1±0.5 &
  81.2±0.2 &
  85.3±0.2 \\
\multicolumn{1}{c|}{} &
  \multicolumn{1}{c|}{MPA} &
  82.5±1.1 &
  82.5±0.2 &
  91.4±0.1 &
  80.2±0.1 &
  \multicolumn{1}{c|}{84.2±0.2} &
  83.5±0.2 &
  82.6±0.2 &
  91.8±0.0 &
  80.4±0.3 &
  84.4±0.1 \\ \hline
\multicolumn{2}{c|}{VAMP} &
  \textbf{84.3}±0.1 &
  \textbf{84.3}±0.1 &
  \textbf{92.6}±0.1 &
  \textbf{80.8}±0.2 &
  \multicolumn{1}{c|}{\textbf{85.5}±0.0} &
  \textbf{85.1}±0.1 &
  \textbf{84.9}±0.1 &
  \textbf{92.5}±0.0 &
  \textbf{81.9}±0.0 &
  \textbf{86.1}±0.0 \\ \hline
\end{tabular}
\caption{ Accuracy (\%) and Standard deviation (\%) evaluation on target domain of OfficeHome and DomainNet dataset. A, P, R, and C in OfficeHome denote acronyms of Art, Product, Real, and Clipart, respectively. P, R, S, and C in DomainNet denote acronyms of Painting, Real, Sketch, and Clipart, respectively. Our reported results are the average of four runnings. The best results are shown in bold. Mann-Whitney U test is performed to compared our average results with the second-best average results. On OfficeHome, p-values are 0.01 for \%3 and 0.01 for \%6. On DomainNet, p-values are 0.02 for 1 shot and 0.01 for 3 shot. The p-values are all less than 0.05 indicating a significant difference in medians.}
\label{tab:tab1}
\end{table*}

\noindent \textbf{Text Semantic Diversity.} Intuitively, text prompts among different domains should be diverse to describe the domain-specific semantic meaning. For example, a customized description for the sketch domain could be ``\texttt{A sketch of a <category> with pencil lines.}'', while the painting domain might be ``\texttt{A painting of a <category> with colorful paint.}''. In our implementation, the domain-agnostic text prompts are fixed as the manual-crafted prompt, i.e., {\texttt{a photo of a <category>.}, and the domain-specific text prompt guided by the corresponding visual prompt is concatenated with it. It is expected that the final text prompts from different source domains are encouraged to be slightly dissimilar to better represent the domain-specific descriptions of diversity. The semantic orthogonal constraint is introduced to ensure dissimilarity as follows, 
\begin{equation}
\small
\begin{array}{l}
{{\cal L}_{TSD}} = \frac{1}{{{\cal C}_M^2 \cdot K}}\sum\limits_{i = 1}^{M - 1} {\sum\limits_{r = i + 1}^M {\sum\limits_{k = 1}^K {\left| {\cos (W_k^i,W_k^r)} \right|} } } \\
 = \frac{1}{{{\cal C}_M^2 \cdot K}}\sum\limits_{i = 1}^{M - 1} {\sum\limits_{r = i + 1}^M {\sum\limits_{k = 1}^K {\left| {\cos ({\Phi _i}({{\bf{E}}_k},{{\bf{P}}_i}),{\Phi _r}({{\bf{E}}_k},{{\bf{P}}_r}))} \right|} } } 
 \end{array}
\label{eq:eq14}
\end{equation}
where ${W^i}$ represents the text representations from the $i$-th source domain.

\noindent \textbf{Training VAMP Framework.}
Accordingly, the total optimization objective within each edge-side model and among edge-side models is,
\begin{equation}
\small
{\cal L} = {{\cal L}_{CSA}} + {\alpha _1} \cdot \lambda ({{\cal L}_{DDA}} + {{\cal L}_{TCC}}) + {\alpha _2} \cdot \lambda {{\cal L}_{TSD}}
\label{eq:eq15}
\end{equation}
where  $\alpha$s are the hyperparameters to balance losses. Considering that each domain-specific prompt is not fully trained at the beginning, $\lambda$ is a dynamically adjustive coefficient to control $\alpha$s, which increases with the training steps and is calculated as,
\begin{equation}
\small
\lambda  = 2 \cdot {\rm{sigmoid}}(10 \cdot \frac{{steps}}{{total\_steps}}) - 1
\label{eq:16}
\end{equation}
The pseudo-training procedure of the VAMP is shown in Algorithm I, in Appendix A. During inference in the centralized integration platform, the average predicted logits from the multiple edge-side devices are used as the final prediction of target samples, which can be quickly achieved by inserting the domain-specific prompts they send.
\section{Experiments}
This section describes the datasets, baselines, extensive experiments, and analysis. More details about datasets and implementations are shown in Appendix B and C.
\subsection{Experimental Settings}
\noindent \textbf{Datasets.} VAMP is evaluated on two multi-source few-shot domain adaptation benchmarks, OfficeHome \cite{Venkateswara2017}, and DomainNet \cite{Peng2019}.

\noindent \textbf{Baselines.} We compare the VAMP with the several prompt tuning methods implemented under the UMFDA scenario and based on ViT-B/16 CLIP, which are organized into the following types.
\begin{itemize}
    \item \textbf{Zero-shot CLIP.} It denotes that CLIP zero-shot inference on the target domain data is directly implemented. 
    \item  \textbf{Domain-agnostic Prompt.} It includes several prevalent prompt tuning methods, such as CoOp \cite{Zhou2022a}, VPT (denotes only vision-branch prompt tuning) and MaPLe \cite{Khattak2023}. In UMFDA, they are viewed as domain-specific prompts that are independently learned on each edge-side model. Only a few annotated data can be used to train. In the central model, the best inference result in the target domain is reported as the final inference result.
    \item \textbf{Disentangling Prompt Tuning.} It includes a single-source domain adaptation method DAPL \cite{Ge2022} and a multi-source domain adaptation method MPA \cite{Chen2023}. Similar to the domain-agnostic prompt method mentioned above, DAPL is independently implemented on each edge-side model, and the best inference results are reported. For MPA, the reported results are from the inference after the second-stage alignment training on the central model.
\end{itemize}
\begin{table}[!t]
\scriptsize
\centering
\setlength{\tabcolsep}{1.0mm} 
\renewcommand{\arraystretch}{0.8}
\begin{tabular}{ccccccccccc}
\hline
\multicolumn{11}{c}{OfficeHome} \\ \hline
\multicolumn{1}{c|}{\multirow{3}{*}{Direction}} &
  \multicolumn{5}{c|}{3\%} &
  \multicolumn{5}{c}{6\%} \\ \cline{2-11} 
\multicolumn{1}{c|}{} &
  \multirow{2}{*}{\begin{tabular}[c]{@{}c@{}}APR\\ →C\end{tabular}} &
  \multirow{2}{*}{\begin{tabular}[c]{@{}c@{}}CPR\\ →A\end{tabular}} &
  \multirow{2}{*}{\begin{tabular}[c]{@{}c@{}}CAR\\ →P\end{tabular}} &
  \multirow{2}{*}{\begin{tabular}[c]{@{}c@{}}CAP\\ →R\end{tabular}} &
  \multicolumn{1}{c|}{\multirow{2}{*}{Avg}} &
  \multirow{2}{*}{\begin{tabular}[c]{@{}c@{}}APR\\ →C\end{tabular}} &
  \multirow{2}{*}{\begin{tabular}[c]{@{}c@{}}CPR\\ →A\end{tabular}} &
  \multirow{2}{*}{\begin{tabular}[c]{@{}c@{}}CAR\\ →P\end{tabular}} &
  \multirow{2}{*}{\begin{tabular}[c]{@{}c@{}}CAP\\ →R\end{tabular}} &
  \multirow{2}{*}{Avg} \\
\multicolumn{1}{c|}{} &
   &
   &
   &
   &
  \multicolumn{1}{c|}{} &
   &
   &
   &
   &
   \\ \hline
\multicolumn{1}{c|}{\textit{vision→text}} &
  \textbf{74.5} &
  \textbf{85.6} &
  \textbf{91.9} &
  90.6 &
  \multicolumn{1}{c|}{\textbf{85.7}} &
  \textbf{73.6} &
  \textbf{86.3} &
  \textbf{91.3} &
  \textbf{91.3} &
  \textbf{85.6} \\
\multicolumn{1}{c|}{\textit{text→vision}} &
  73.6 &
  85.4 &
  91.1 &
  \textbf{90.7} &
  \multicolumn{1}{c|}{85.2} &
  73.3 &
  85.6 &
  91.0 &
  91.0 &
  85.2 \\ \hline
\multicolumn{11}{c}{DomainNet} \\ \hline
\multicolumn{1}{c|}{\multirow{3}{*}{Direction}} &
\multicolumn{5}{c|}{1 shot} &
  \multicolumn{5}{c}{3 shot} \\ \cline{2-11}
\multicolumn{1}{c|}{} &
  \multirow{2}{*}{\begin{tabular}[c]{@{}c@{}}PRS\\ →C\end{tabular}} &
  \multirow{2}{*}{\begin{tabular}[c]{@{}c@{}}CRS\\ →P\end{tabular}} &
  \multirow{2}{*}{\begin{tabular}[c]{@{}c@{}}CPS\\ →R\end{tabular}} &
  \multirow{2}{*}{\begin{tabular}[c]{@{}c@{}}CPR\\ →S\end{tabular}} &
  \multicolumn{1}{c|}{\multirow{2}{*}{Avg}} &
  \multirow{2}{*}{\begin{tabular}[c]{@{}c@{}}PRS\\ →C\end{tabular}} &
  \multirow{2}{*}{\begin{tabular}[c]{@{}c@{}}CRS\\ →P\end{tabular}} &
  \multirow{2}{*}{\begin{tabular}[c]{@{}c@{}}CPS\\ →R\end{tabular}} &
  \multirow{2}{*}{\begin{tabular}[c]{@{}c@{}}CPR\\ →S\end{tabular}} &
  \multirow{2}{*}{Avg} \\
\multicolumn{1}{c|}{} &
   &
   &
   &
   &
  \multicolumn{1}{c|}{} &
   &
   &
   &
   &
   \\ \hline
\multicolumn{1}{c|}{\textit{vision→text}} &
  \textbf{84.3} &
  \textbf{84.3} &
  \textbf{92.6} &
  \textbf{80.9} &
  \multicolumn{1}{c|}{\textbf{85.5}} &
  \textbf{85.0} &
  \textbf{84.9} &
  \textbf{92.5} &
  \textbf{81.9} &
  \textbf{86.1} \\
\multicolumn{1}{c|}{\textit{text→vision}} &
  83.4 &
  83.7 &
  92.2 &
  80.8 &
  \multicolumn{1}{c|}{85.0} &
  84.9 &
  84.9 &
  92.3 &
  81.8 &
  86.0 \\ \hline
\end{tabular}
\caption{Ablation studies on changing the direction of prompt projection in the proposed framework.}
\label{tab:tab2}
\end{table}
\subsection{Comparative Results}
\begin{table}[!t]
\centering
\scriptsize
\setlength{\tabcolsep}{0.6mm} 
\renewcommand{\arraystretch}{0.8}
\begin{tabular}{ccc|ccccc|ccccc}
\hline
\multirow{3}{*}{$TSD$} &
  \multirow{3}{*}{$TCC$} &
  \multirow{3}{*}{$DDA$} &
  \multicolumn{5}{c|}{3\%} &
  \multicolumn{5}{c}{6\%} \\ \cline{4-13} 
 &
   &
   &
  \multirow{2}{*}{\begin{tabular}[c]{@{}c@{}}APR\\ →C\end{tabular}} &
  \multirow{2}{*}{\begin{tabular}[c]{@{}c@{}}CPR\\ →A\end{tabular}} &
  \multirow{2}{*}{\begin{tabular}[c]{@{}c@{}}CAR\\ →P\end{tabular}} &
  \multirow{2}{*}{\begin{tabular}[c]{@{}c@{}}CAP\\ →R\end{tabular}} &
  \multirow{2}{*}{Avg} &
  \multirow{2}{*}{\begin{tabular}[c]{@{}c@{}}APR\\ →C\end{tabular}} &
  \multirow{2}{*}{\begin{tabular}[c]{@{}c@{}}CPR\\ →A\end{tabular}} &
  \multirow{2}{*}{\begin{tabular}[c]{@{}c@{}}CAR\\ →P\end{tabular}} &
  \multirow{2}{*}{\begin{tabular}[c]{@{}c@{}}CAP\\ →R\end{tabular}} &
  \multirow{2}{*}{Avg} \\
 &
   &
   &
   &
   &
   &
   &
   &
   &
   &
   &
   &
   \\ \hline
 &
   &
   &
  \textbf{74.5} &
  85.6 &
  \textbf{91.9} &
  90.6 &
  \textbf{85.7} &
  \textbf{73.6} &
  \textbf{86.3} &
  \textbf{91.3} &
  91.3 &
  \textbf{85.6} \\
$\scriptsize \usym{1F5F4}$ &
   &
   &
  73.7 &
  \textbf{85.7} &
  91.7 &
  \textbf{90.8} &
  85.5 &
  73.4 &
  85.6 &
  91.3 &
  \textbf{91.5} &
  85.5 \\
$\scriptsize \usym{1F5F4}$ &
  $\scriptsize \usym{1F5F4}$ &
   &
  73.7 &
  85.7 &
  91.4 &
  90.3 &
  85.3 &
  73.4 &
  85.7 &
  91.2 &
  91.1 &
  85.4 \\
$\scriptsize \usym{1F5F4}$ &
  $\scriptsize \usym{1F5F4}$ &
  $\scriptsize \usym{1F5F4}$ &
  74.0 &
  85.2 &
  91.7 &
  90.3 &
  85.3 &
  73.5 &
  85.9 &
  91.2 &
  91.1 &
  85.4 \\ \hline
\end{tabular}
\caption{Ablation studies on various losses of VAMP on the OfficeHome dataset.}
\label{tab:tab3}
\end{table}
Extensive experiments are conducted on OfficeHome and DomainNet datasets, as shown in Table \ref{tab:tab1}. The observation and findings are as follows. (1) The zero-shot CLIP inference has achieved a good performance, which demonstrates that CLIP has the advantage of language supervision and the original text prompt is capable of inspiring the generalization knowledge of CLIP. (2) Within the domain-agnostic prompt tuning methods, MaPLe works best, demonstrating the advantage of multimodal prompt tuning to maintain the discriminability of features. (3) In the UMFDA setting, DAPL's and MPA's performances 
do not meet expectations. Their results reflect that decoupling prompts to capture domain-specific and domain-agnostic contexts by contrastive learning are suboptimal under inadequately annotated source domain data. (4) VAMP achieves competitive performance compared to baselines and has an average improvement of 3.2\% and 1.6\% compared to the zero-shot CLIP inference on the OfficeHome and DomainNet datasets, respectively. It indicates that utilizing vision-aware multimodal prompts as domain-specific prompts to perceive domain information alleviates the limitations of disentangling prompts in the few-shot setting. Further, it demonstrates that this decentralized training framework of the VAMP can simultaneously address the issues of domain and semantic alignment within edge-side models and collaborative learning among edge-side models without central training.
\begin{table}[!t]
\centering
\scriptsize
\setlength{\tabcolsep}{1.0mm} 
\renewcommand{\arraystretch}{0.8}
\begin{tabular}{c|ccccc|ccccc}
\hline
\multirow{3}{*}{b} & \multicolumn{5}{c|}{3\%}         & \multicolumn{5}{c}{6\%}          \\ \cline{2-11} 
 &
  \multirow{2}{*}{\begin{tabular}[c]{@{}c@{}}APR\\ →C\end{tabular}} &
  \multirow{2}{*}{\begin{tabular}[c]{@{}c@{}}CPR\\ →A\end{tabular}} &
  \multirow{2}{*}{\begin{tabular}[c]{@{}c@{}}CAR\\ →P\end{tabular}} &
  \multirow{2}{*}{\begin{tabular}[c]{@{}c@{}}CAP\\ →R\end{tabular}} &
  \multirow{2}{*}{Avg} &
  \multirow{2}{*}{\begin{tabular}[c]{@{}c@{}}APR\\ →C\end{tabular}} &
  \multirow{2}{*}{\begin{tabular}[c]{@{}c@{}}CPR\\ →A\end{tabular}} &
  \multirow{2}{*}{\begin{tabular}[c]{@{}c@{}}CAR\\ →P\end{tabular}} &
  \multirow{2}{*}{\begin{tabular}[c]{@{}c@{}}CAP\\ →R\end{tabular}} &
  \multirow{2}{*}{Avg} \\
                   &      &      &      &      &      &      &      &      &      &      \\ \hline
2                  & 73.2 & 85.6 & 91.5 & 91.0 & 85.3 & 72.7 & 85.5 & 91.5 & 91.6 & 85.3 \\
4                  & 73.2 & 85.2 & 91.4 & 91.0 & 85.2 & 73.4 & 85.8 & 91.5 & 91.4 & 85.5 \\
8                  & 73.4 & 85.5 & 91.6 & 91.2 & 85.4 & 72.9 & 86.1 & 91.4 & 91.4 & 85.5 \\
16                 & 73.7 & 85.9 & 91.0 & 90.9 & 85.4 & 73.5 & 85.5 & 91.5 & 91.4 & 85.5 \\ \hline
\end{tabular}
\caption{Ablation studies on prompt lengths $b$ of VAMP on OfficeHome dataset.}
\label{tab:tab4}
\end{table}
\begin{figure}[!t]
    \centering
    \captionsetup{skip=2pt}  
    \begin{minipage}{\linewidth}
    \centering
       \subfigure[\textit{text→vision}]{
        \includegraphics[width=1.0\linewidth]{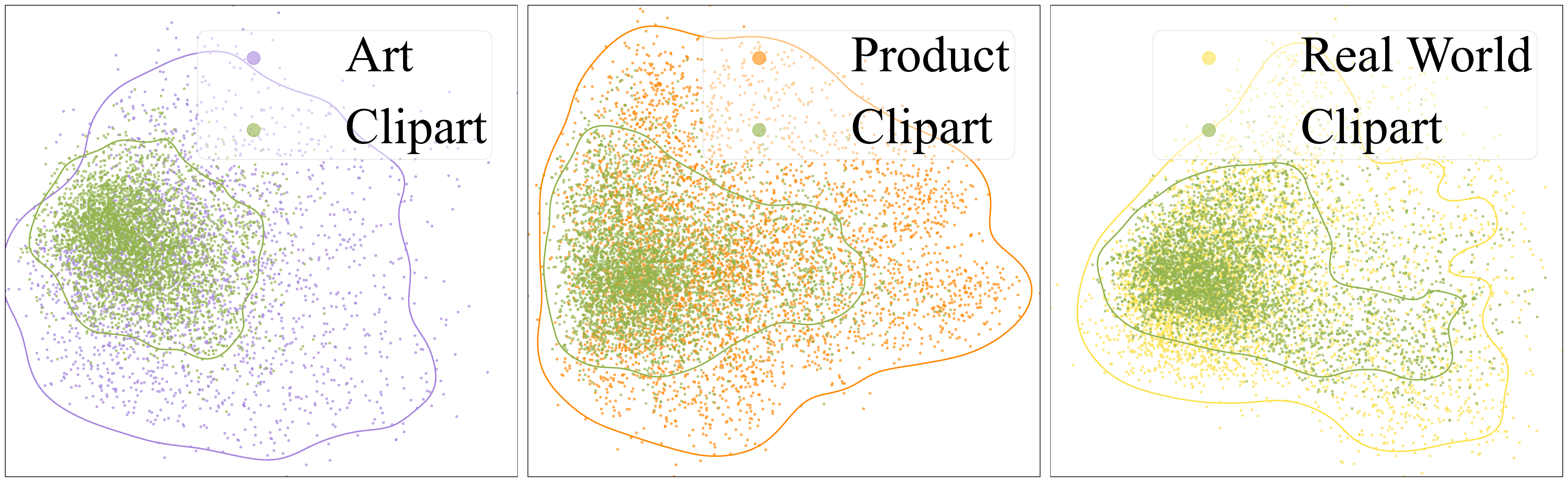}
        }
    \end{minipage}
    \begin{minipage}{\linewidth}
        \centering
        \subfigure[\textit{vision→text}]{
        \includegraphics[width=1.0\linewidth]{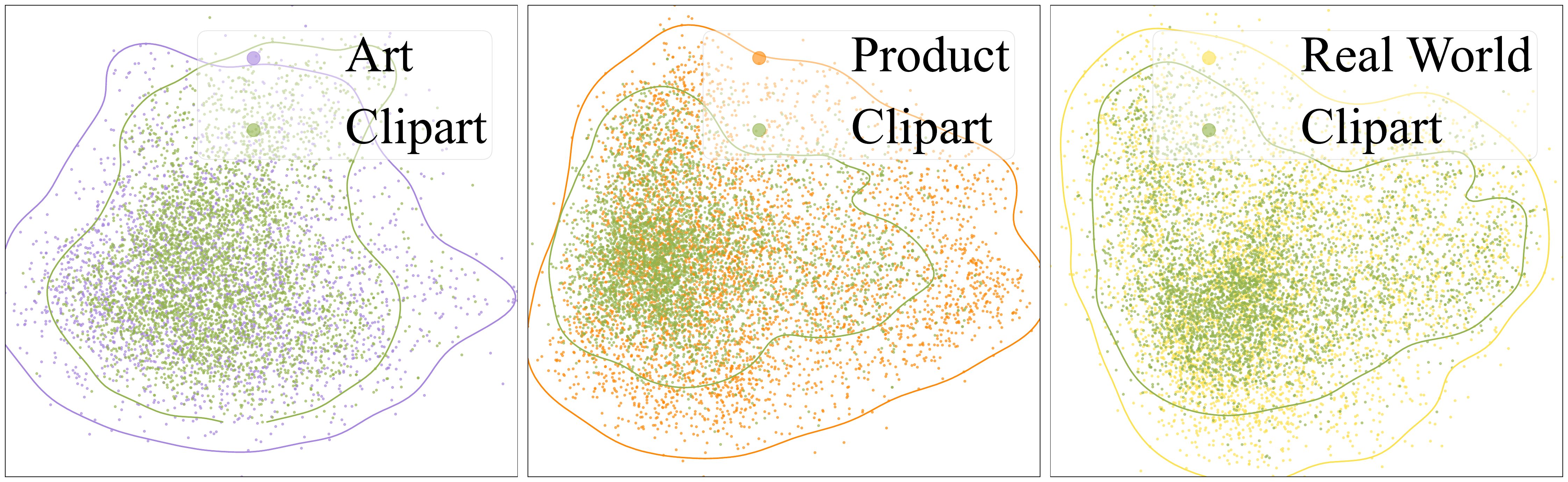}
        }
    \end{minipage}
    \caption{PCA visualizations of the extracted image features from the source and target domains in different domain-specific models of "APR→C". The first row of pictures denotes that the projection direction of prompts is from \textit{text} to \textit{vision}; the second row represents the \textit{vision}-to-\textit{text} projection that is used in the VAMP.  Contour lines enclose the regions with high density of data points.}
    \label{fig:fig4}
\end{figure}
\subsection{Ablation Studies}
\noindent \textbf{Effects of Vision-aware Multimodal Prompts.} Table \ref{tab:tab2} shows the effectiveness of vision-aware multimodal prompts in projecting direction from vision to text (\textit{vision→text}). It demonstrates that this multimodal prompt tuning is better for perceiving domain information from the vision encoder and affects the text prompt tuning.

\noindent \textbf{Effects of Various Losses.} Table \ref{tab:tab3} shows the declines of various degrees for the VAMP framework when various losses are removed on OfficeHome. It validates the effectiveness of these introduced losses to jointly optimize the VAMP better.

\noindent \textbf{Effects of Prompt Length.} Table \ref{tab:tab4} shows the effects of prompt length for the vision-aware multimodal prompts on OfficeHome. The ablation results indicate that increasing its prompt length has a limited impact on performance.
\subsection{Quantity Analysis}
\noindent \textbf{Domain Distribution Visualization.} Figure \ref{fig:fig4} depicts the domain distribution visualization in two projection directions (\textit{text→vision} and \textit{vision→text}). The \textit{vision→text} projection has a better alignment between the source domain and target domain in each domain-specific image extractor, demonstrating the effectiveness of vision-aware multimodal prompt to perceive the domain information.

\noindent \textbf{t-SNE Visualization and Variance Statistics.} Figure \ref{fig:fig5} shows the different t-SNE visualization and variance statistics of the image and text features from target domain. The observation is that the image features of VAMP are more concentrated aggregations within classes, and it has fewer intra-class visual variances. The text feature distribution of VAMP between classes is not even as DAPL. Because of only text-branch prompt tuning, DAPL has a more significant inter-class text variance, showing more explicit inter-class boundaries. Although there is a loss in the inter-class distribution of text features, it demonstrates that our VAMP can balance feature discriminability and domain alignment by vision-aware prompt influencing text prompt. Case studies about visualizing attention maps are presented in Appendix D.
\section{Related Work}
\noindent \textbf{MFDA.} MFDA is a more common setting \cite{Gulshan2016,Harmon2020} since it only has fewer annotated source samples. The first insight into MFDA comes from MSFAN \cite{Yue2021a}. Like most conventional multi-source domain adaptation (MDA) methods \cite{Venkat2020,Peng2019,Xu2018}, its framework comprises a shared feature and multiple domain-specific classifiers. MSFAN adopts similarity-based classifiers \cite{Saito2019}, which are trained by prototypical contrastive learning \cite{Yue2021b} in each pair of source and target domains to learn the well-clustered prototype features with better semantic discriminability, instead of using the linear classifiers. This paper further considers a UMFDA setting suited for the resource-limited edge device. More comparisons of our VAMP in the MFDA are presented in Appendix E.
\begin{figure}[!t]
    \centering
    \captionsetup{skip=1pt}
    \subfigure[VAMP]{\includegraphics[width=0.49\linewidth]{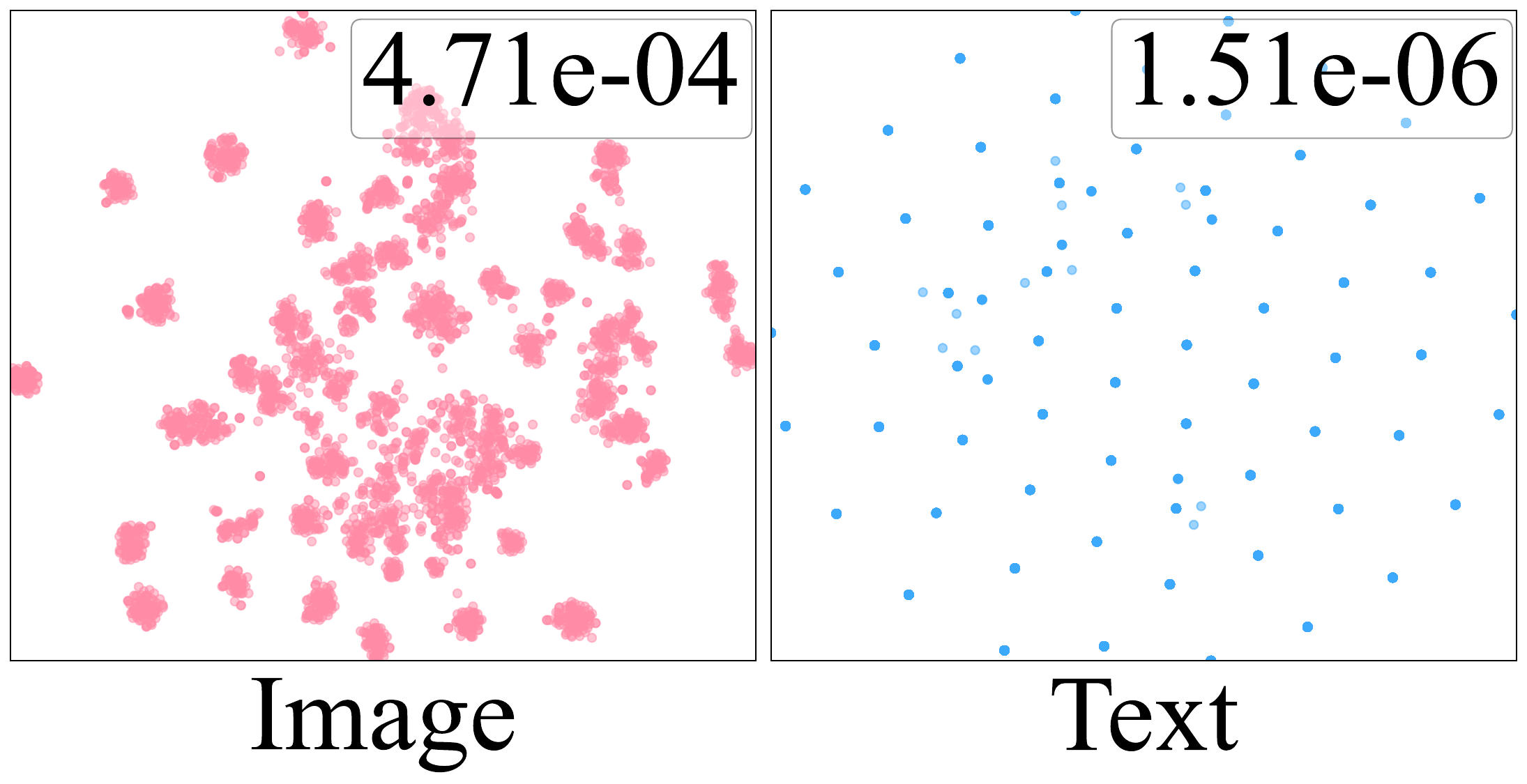}}
    \subfigure[DAPL]{\includegraphics[width=0.49\linewidth]
    {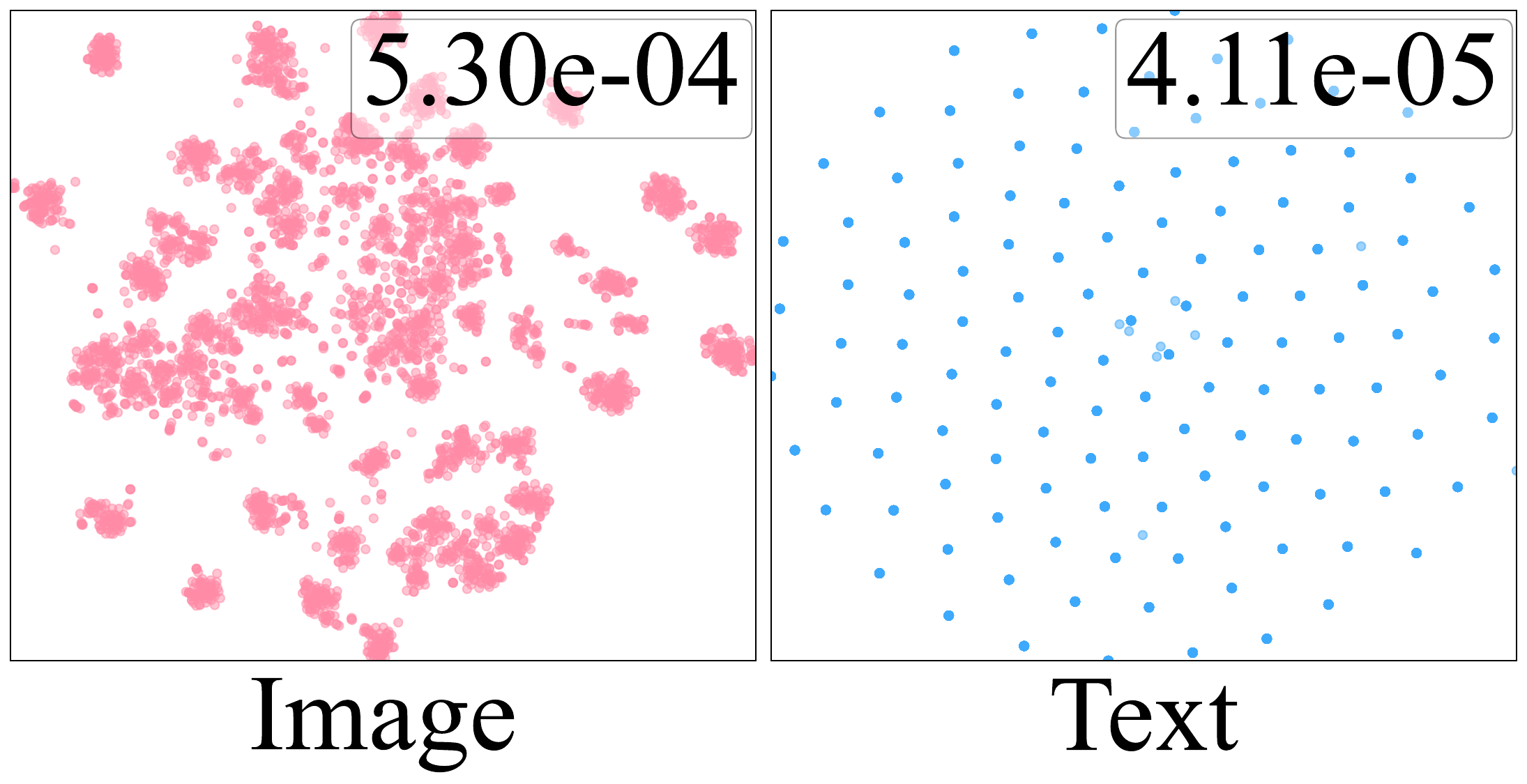}}
    \caption{t-SNE visualization of the image and text features of target domain extracted by "Clipart-Real World" model of VAMP and DAPL. The statistics of either intra-class visual variance or inter-class text variance are shown at the top of the subfigure.}
    \label{fig:fig5}
\end{figure}

\noindent \textbf{Prompt Tuning in Domain Adaptation.}
Prompt tuning technology has efficiently adapted pretrained models to various domains \cite{Jia2022, Yao2021,Jin2022,Ju2022}. Some domain adaptation prompt tuning methods have emerged recently \cite{Ge2022, Chen2023,Singha2023,Bai2024,Wang2024}. DAPL \cite{Ge2022} pioneers the application of prompt tuning in single-source unsupervised domain adaptation. It disentangles context prompts as domain-agnostic and domain-specific prompts, where the domain information is shared by the same domain and thus dynamically adapts to the text classifier encoder. Subsequently, MPA \cite{Chen2023} continually develops the idea of disentangling prompts in the MDA. After learning individual prompts for each pair of source and target domains, they excavate the relations among the learned prompts by the auto-encoder networks. Differently, this paper explores the prompt tuning in MDA under a resource-limited scenario with limited data.

\section{Conclusions}
This study introduces a UMFDA schema for the resource-limited edge computing scenario, inspired by the advantage of pretrained VLM and the plug-and-play prompt tuning capable of efficiently transferring the VLM. We further propose a decentralized training VAMP framework. It is based on the customized vision-aware multimodal prompts to perceive the domain information and maintain the features' discriminability. Within the VAMP framework, CSA, DDA, TCC, and TSD losses are introduced to optimize the alignment inside the edge-side model and to enhance collaboration among edge-side models. The experimental results demonstrate the VAMP's effectiveness. Future work will explore VAMP's performance on more edge-side models in the various downstream tasks of MFDA. 

\section{Acknowledgements}
This work was supported by the National Natural Science Foundation of China (NSFC) under Grant Nos.61966038, 62266051, 62202416 and 62162068, and the Postgraduate Research and Innovation Foundation of Yunnan University under Grant No.KC-24248816. The authors would like to thank the anonymous reviewers for their constructive comments.

\bibliography{MyCollection}
\newpage
\section{Appendix}
The appendix contains supplementary materials that provide additional details for the main paper and more experimental analysis, which can be organized as follows:
\begin{itemize}
    \item Section A provides the pseudo-training procedure of VAMP.
    \item Section B provides the details and statistics for datasets.
    \item Section C describes the implementations including VAMP and other baselines.
    \item Section D presents the case studies that visualize the attention maps from images in the target domain. 
    \item Section E shows that VAMP under the MFDA setting is compared with the conventional methods.
\end{itemize}

\subsection{A. Algorithm of VAMP}
The pseudo-training procedure of the VAMP framework is shown as Algorithm \ref{alg:algorithm}.
\subsection{B. Dataset Details}
For OfficeHome \cite{Venkateswara2017}, and DomainNet \cite{Peng2019} datasets, the annotated data in each source domain follows splits of PCS \cite{Yue2021b}, CDS \cite{Kim2020} and MSFAN \cite{Yue2021a}, where the amount of annotated data is much smaller than the unannotated. Each domain is regarded as the target domain in turn; the others become the source domains. 
Specifically, OfficeHome has 65 categories in four domains: Art, Clipart, Product, and Real. Following MFSAN and PCS's setting, experiments are conducted with 3\% and 6\% labeled source images per category, so the number of categories is unbalanced. DomainNet is a large-scale domain adaptation dataset. We follow the previous research \cite{Kim2020,Yue2021a,Yue2021b} using a subset containing four domains (Clipart, Painting, Real, Sketch) with 126 categories, because of the noise effect. 1-shot and 3-shot source label data in this dataset are utilized. Their statistics and the number of annotated samples in each domain are shown in Table. \ref{tab:tab5}.
\begin{algorithm}[!t]
\caption{The pseudo-training procedure of VAMP.}
\label{alg:algorithm}
\begin{algorithmic}[1]
    \STATE \textbf{Input:} source domain datasets $\{ D_s^i = D_{s\_a}^i \cup D_{s\_u}^i\} _{i = 1}^M$; target domain dataset $D_t$; a group of domain-specific vision-aware multimodal prompts $\{ {{\bf{P}}_1},{{\bf{P}}_2},...,{{\bf{P}}_M}\}$ for $M$ edge-side models; the frozen image encoder $\Psi ( \cdot )$ and text encoder $\Phi ( \cdot )$ of CLIP on each edge device; the fixed token embeddings of all categories ${\bf{E}}$; pseudo label threshold ${\cal T}$; a dynamically adjustive coefficient $\lambda$; training $T$ iterations.
    \STATE \textbf{Output:} The trained $\{ {{\bf{P}}_1},{{\bf{P}}_2},...,{{\bf{P}}_M}\}$.
    \STATE Construct $ M$ the domain-specific image feature extractors $\{ {\Psi _i}( \cdot ,{{\bf{P}}_i})\} _{i = 1}^M$ and text classifiers $\{ {\Phi _i}({\bf{E}},{{\bf{P}}_i})\} _{i = 1}^M$ for edge-side models by inserting $\{ {{\bf{P}}_1},{{\bf{P}}_2},...,{{\bf{P}}_M}\}$.
    \FOR{$t=0$ to $T$} 
        \STATE Randomly choose an edge-side model $\rm i$ from $M$ edge-side models, and obtain its corresponding $D_{s}^{\rm i}$.
        \STATE Randomly sample a batch of annotated data $\{ (x_j^{a,{\rm i}},y_j^{a,{\rm i}})\} _{j = 1}^{n_a^{\rm i}}$  from $D_{s\_a}^{\rm i}$ and unannotated data $\{ x_j^{u,{\rm i}}\} _{j = 1}^{n_u^{\rm i}}$  from $D_{s\_u}^{\rm i}$ on the edge-side model ${\rm i}$.
        \STATE Randomly sample a batch of target data $\{ x_j^t\} _{j = 1}^{{n_t}}$ from $D_t$ send to $M$ edge side models. \\
        \mycomment{\# Training in the edge-side model ${\rm i}$.}
        \STATE Align cross-modal semantic information using $\{ (x_j^{a,{\rm i}},y_j^{a,{\rm i}})\} _{j = 1}^{n_a^{\rm i}}$, $\{ x_j^t\} _{j = 1}^{{n_t}}$, ${\Psi _{\rm i}}( \cdot ,{{\bf{P}}_{\rm i}})$ and ${\Phi _{\rm i}}({\bf{E}},{{\bf{P}}_{\rm i}})$ by Eq.(9), where the pseudo labels of $\{ x_j^t\} _{j = 1}^{{n_t}}$ are obtained by zero-shot CLIP inference under the certain condition of ${\cal T}$.
        \STATE Align distribution between the pair of the source domain and target domain using $\{ x_j^{u,{\rm i}}\} _{j = 1}^{n_u^{\rm i}}$, $\{ x_j^t\} _{j = 1}^{{n_t}}$ and ${\Psi _{\rm i}}( \cdot ,{{\bf{P}}_{\rm i}})$ by Eq.(12). \\
        \mycomment{\# Collaborative training among $M$ edge-side models.}
        \STATE Keep all text classifiers' outputs consistent using $\{ x_j^t\} _{j = 1}^{{n_t}}$, $\{ {\Psi _i}( \cdot ,{{\bf{P}}_i})\} _{i = 1}^M$ and $\{ {\Phi _i}({\bf{E}},{{\bf{P}}_i})\} _{i = 1}^M$ by Eq.(13).
        \STATE Encourage the semantic diversity using $\{ {\Phi _i}({\bf{E}},{{\bf{P}}_i})\} _{i = 1}^M$ by Eq.(14).
        \STATE Update $\lambda$ according to $t$ and $T$ by Eq.(16).
        \STATE $\{ {{\bf{P}}_1},{{\bf{P}}_2},...,{{\bf{P}}_M}\}$ is optimized by minimizing Eq. (15).
    \ENDFOR
\end{algorithmic}
\end{algorithm}
\subsection{C. Implementation Details}
\noindent \textbf{Implementation Details of VAMP.} The experiments adopt the ViT-B/16 CLIP backbone for vision-aware multimodal prompt tuning, so $d_T = 512$, $d_V = 768$ and  $d = 512$. The parameters of the CLIP encoder remain frozen. The prompt inserted depth $J$ is 12. The prompt length $b$ is 16 for OfficeHome and 8 for DomainNet, respectively. In Eq.(15), set ${\alpha}_1$ to 0.1 and ${\alpha}_2$ to 0.01, respectively. VAMP is trained with 50 and 20 epochs in OfficeHome and DomainNet, respectively, with an annotated source batch size of 4 and an unannotated target and source samples batch size of 64, where the iteration steps of one epoch depend on the number of the annotated source data. The learning rate of VAMP via SGD optimizer is 0.003. The threshold ${\cal T}$ is set to 0.8. All the experiments are conducted on a single NVIDIA 3090 GPU. All reported results are the averages of 4 training runs with different random seeds.

\noindent \textbf{Implementation Details of Baselines.} Our implementation details of baselines are shown as follows, where the settings for CoOp and VPT (denotes only vision branch prompt tuning) refer to the few-shot setting in the paper of MaPLe \cite{Khattak2023}. They all use ViT-B/16 CLIP as the backbone and are trained under 4 random seeds.
\begin{itemize}
    \item \textbf{CoOp.} It was trained for 10 epochs with a batch size of 4 and a learning rate of 0.0035 via SGD optimizer.
    \item \textbf{VPT.} It used the deep variant of vision prompt tuning, where the prompt depth was 12 and the prompt length was 4 in the vision branch. It was trained with a batch size of 4, a learning rate of 0.0025, and an epoch of 10. 
    \item \textbf{MaPLe.} It set the prompt depth to 9 and the prompt lengths to 2 for both text and vision encoder. The model was trained for 5 epochs with a batch size of 4 and a learning rate 0.0035 via SGD optimizer.
    \item \textbf{DAPL.} The domain-agnostic and domain-specific prompt lengths were set to 16. The class-specific context is not set. The pseudo target labeling threshold was set as 0.8, the same as ours. The batch sizes for annotated source samples and unannotated target samples were 32. It was trained 25 epochs with a learning rate of 0.003 via SGD optimizer.
    \item \textbf{MPA.} The domain-invariant and domain-specific prompt lengths were also set to 16 and 16 for the hyper-parameters, respectively. The class-specific context is also not set. Likewise, the target pseudo-label threshold was fixed at 0.8. The dimensions of the intrinsic subspace are 150 and 250 for OfficeHome and DomainNet, respectively. The prompts and auto-encoder were trained with a learning rate of 0.003 and 0.005 via SGD optimizer. The training iterations of the first prompt learning stage and the second multi-prompt alignment stage were set to 200 and 100 for OfficeHome and 300 and 300 for DomianNet.
\end{itemize}
\begin{table}[t]
\centering
\resizebox{\linewidth}{!}{
\begin{tabular}{c|c|ccc|c}
\hline
\multirow{2}{*}{Dataset}    & \multirow{2}{*}{Domain} & \multicolumn{3}{c|}{\# Image}  & \multirow{2}{*}{\# Classes}  \\ \cline{3-5}
 &              & \# 3\%    & \# 6\%    & \# total &  \\ \hline
\multirow{4}{*}{OfficeHome} & Art (A)                 & 73       & 146    & 2,427     & \multirow{4}{*}{65}         \\
 & Clipart (C)  & 131      & 262      & 4,365   &  \\
 & Product (P)  & 133      & 266      & 4,439   &  \\
 & Real (R)     & 131      & 261      & 4,357   &  \\ \hline
\multirow{2}{*}{Dataset}    & \multirow{2}{*}{Domain} & \multicolumn{3}{c|}{\# Image} & \multirow{2}{*}{\# Classes} \\ \cline{3-5}
 &              & \# 1 shot & \# 3 shot & \# total &  \\ \hline
\multirow{4}{*}{DomainNet}  & Clipart (C)             & 126      & 378    & 18,703    & \multirow{4}{*}{126}        \\
 & Painting (P) & 126      & 378      & 31,502  &  \\
 & Real (R)     & 126      & 378      & 70,358  &  \\
 & Sketch (S)   & 126      & 378      & 24,582  &  \\ \hline
\end{tabular}
}
\caption{Statistics information of OfficeHome and DomainNet datasets.}
\label{tab:tab5}
\end{table}
\subsection{D. Case studies}
Figure \ref{fig:fig6} visualizes the attention map of zero-shot CLIP, MaPLe and the proposed VAMP, where the first two rows of images are from the Real Word domain of OfficeHome; the last two rows are from the Sketch domain of DomainNet. The attention score is from the last block of the CLIP-ViT encoder from one domain-specific model (Clipart-Real Word in OfficeHome and Painting-Sketch in DomainNet). Since the DAPL method does not tune the vision encoder, its visualizations are consistent with zero-shot CLIP. It can be observed that the VAMP allows key objects in visual features in the target domain to get more attention.
\subsection{E. Comparative Results in the MFDA Setting}
Our VAMP is also compared with the existing conventional methods in the MFDA setting, where all the reported results of these conventional baselines are re-trained by MSFAN \cite{Yue2021a} in the MFDA. Notably, these conventional baselines usually use ResNet as the backbone. Besides, other prompt tuning methods as a way of efficient tuning are also experimented with in the MFDA scenario. All of them in the MFDA setting can be summarized as follows.
\begin{figure}[!t]
    \centering
    \includegraphics[width=1.0\linewidth]{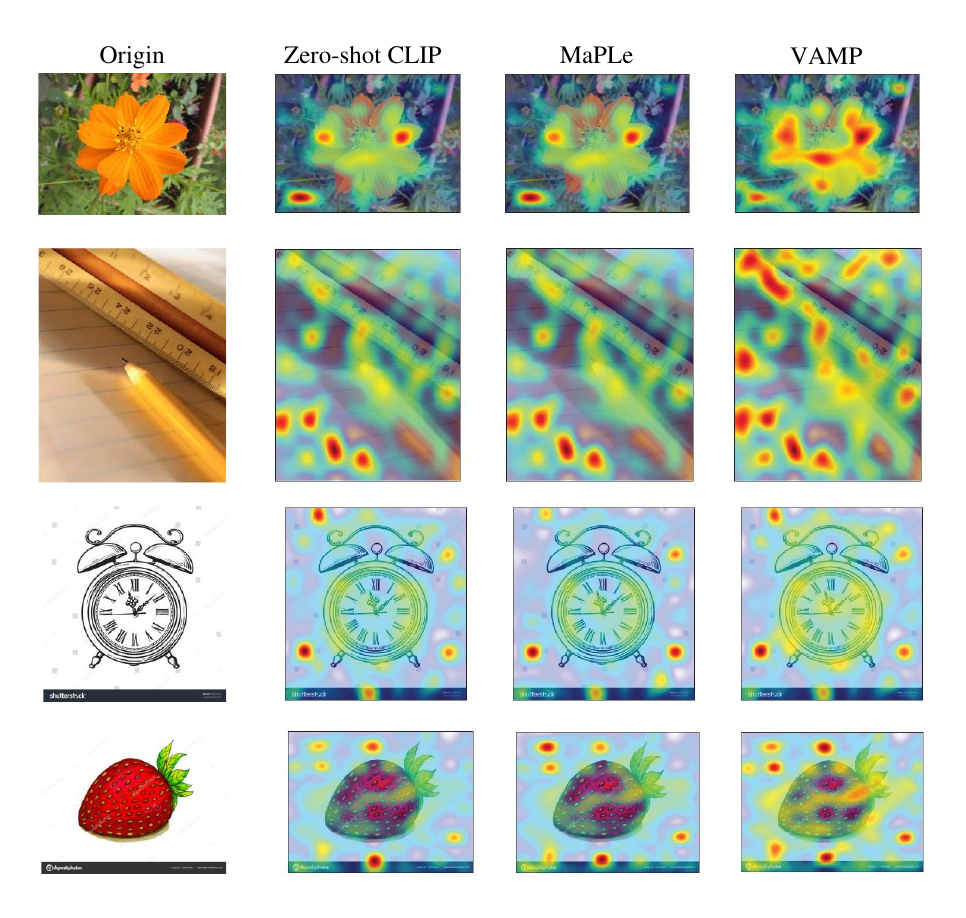}
    \caption{Attention map visualization of zero-shot CLIP, MaPLe and the proposed VAMP.}
    \label{fig:fig6}
\end{figure}
\begin{table*}[!t]
\centering
\small
\resizebox{\linewidth}{!}{
\begin{tabular}{ccccccccccccc}
\hline
\multicolumn{13}{c}{OfficeHome} \\ \hline
\multicolumn{1}{c|}{\multirow{3}{*}{Backbone}} &
  \multicolumn{2}{c|}{\multirow{3}{*}{Method}} &
  \multicolumn{5}{c|}{3\%} &
  \multicolumn{5}{c}{6\%} \\ \cline{4-13} 
\multicolumn{1}{c|}{} &
  \multicolumn{2}{c|}{} &
  \multirow{2}{*}{\begin{tabular}[c]{@{}c@{}}APR\\ →C\end{tabular}} &
  \multirow{2}{*}{\begin{tabular}[c]{@{}c@{}}CPR\\ →A\end{tabular}} &
  \multirow{2}{*}{\begin{tabular}[c]{@{}c@{}}CAR\\ →P\end{tabular}} &
  \multirow{2}{*}{\begin{tabular}[c]{@{}c@{}}CAP\\ →R\end{tabular}} &
  \multicolumn{1}{c|}{\multirow{2}{*}{Avg}} &
  \multirow{2}{*}{\begin{tabular}[c]{@{}c@{}}APR\\ →C\end{tabular}} &
  \multirow{2}{*}{\begin{tabular}[c]{@{}c@{}}CPR\\ →A\end{tabular}} &
  \multirow{2}{*}{\begin{tabular}[c]{@{}c@{}}CAR\\ →P\end{tabular}} &
  \multirow{2}{*}{\begin{tabular}[c]{@{}c@{}}CAP\\ →R\end{tabular}} &
  \multirow{2}{*}{Avg} \\
\multicolumn{1}{c|}{} &
  \multicolumn{2}{c|}{} &
   &
   &
   &
   &
  \multicolumn{1}{c|}{} &
   &
   &
   &
   &
   \\ \hline
\multicolumn{1}{c|}{\multirow{16}{*}{ResNet-50}} &
  \multicolumn{2}{c|}{Single-best} &
  29.0 &
  41.2 &
  52.3 &
  43.1 &
  \multicolumn{1}{c|}{41.4} &
  36.0 &
  49.9 &
  61.8 &
  54.6 &
  50.6 \\ \cline{2-13} 
\multicolumn{1}{c|}{} &
  \multicolumn{2}{c|}{Source-combined} &
  42.2 &
  55.3 &
  63.6 &
  64.1 &
  \multicolumn{1}{c|}{56.3} &
  45.3 &
  60.4 &
  70.5 &
  70.9 &
  61.8 \\ \cline{2-13} 
\multicolumn{1}{c|}{} &
  \multicolumn{1}{c|}{\multirow{5}{*}{Single-best DA}} &
  \multicolumn{1}{c|}{CDAN} &
  27.0 &
  38.7 &
  44.9 &
  40.3 &
  \multicolumn{1}{c|}{37.7} &
  40.1 &
  54.9 &
  63.6 &
  59.3 &
  54.5 \\
\multicolumn{1}{c|}{} &
  \multicolumn{1}{c|}{} &
  \multicolumn{1}{c|}{MME} &
  29.0 &
  39.3 &
  52.0 &
  44.9 &
  \multicolumn{1}{c|}{41.3} &
  37.3 &
  54.9 &
  66.8 &
  61.3 &
  55.1 \\
\multicolumn{1}{c|}{} &
  \multicolumn{1}{c|}{} &
  \multicolumn{1}{c|}{MDDIA} &
  29.5 &
  47.1 &
  56.4 &
  51.0 &
  \multicolumn{1}{c|}{46.0} &
  37.1 &
  58.2 &
  68.4 &
  64.5 &
  57.1 \\
\multicolumn{1}{c|}{} &
  \multicolumn{1}{c|}{} &
  \multicolumn{1}{c|}{CDS} &
  37.8 &
  51.6 &
  53.8 &
  51.0 &
  \multicolumn{1}{c|}{48.6} &
  45.3 &
  63.7 &
  68.6 &
  65.2 &
  60.7 \\
\multicolumn{1}{c|}{} &
  \multicolumn{1}{c|}{} &
  \multicolumn{1}{c|}{PCS} &
  52.5 &
  66.0 &
  75.6 &
  73.9 &
  \multicolumn{1}{c|}{67.0} &
  54.7 &
  67.0 &
  76.6 &
  75.2 &
  68.4 \\ \cline{2-13} 
\multicolumn{1}{c|}{} &
  \multicolumn{1}{c|}{\multirow{5}{*}{Source-combined DA}} &
  \multicolumn{1}{c|}{CDAN} &
  42.6 &
  52.3 &
  64.5 &
  63.2 &
  \multicolumn{1}{c|}{55.7} &
  51.1 &
  67.0 &
  74.2 &
  73.3 &
  66.4 \\
\multicolumn{1}{c|}{} &
  \multicolumn{1}{c|}{} &
  \multicolumn{1}{c|}{MME} &
  42.5 &
  55.4 &
  67.4 &
  64.5 &
  \multicolumn{1}{c|}{57.5} &
  46.0 &
  67.1 &
  75.5 &
  75.7 &
  66.1 \\
\multicolumn{1}{c|}{} &
  \multicolumn{1}{c|}{} &
  \multicolumn{1}{c|}{MDDIA} &
  55.3 &
  66.9 &
  72.3 &
  75.3 &
  \multicolumn{1}{c|}{67.5} &
  57.3 &
  67.2 &
  79.0 &
  74.4 &
  69.5 \\
\multicolumn{1}{c|}{} &
  \multicolumn{1}{c|}{} &
  \multicolumn{1}{c|}{CDS} &
  54.9 &
  66.2 &
  71.6 &
  73.4 &
  \multicolumn{1}{c|}{66.5} &
  54.9 &
  67.5 &
  76.1 &
  77.5 &
  69.0 \\
\multicolumn{1}{c|}{} &
  \multicolumn{1}{c|}{} &
  \multicolumn{1}{c|}{PCS} &
  49.4 &
  67.0 &
  75.0 &
  76.3 &
  \multicolumn{1}{c|}{66.9} &
  50.4 &
  67.0 &
  77.8 &
  79.4 &
  68.7 \\ \cline{2-13} 
\multicolumn{1}{c|}{} &
  \multicolumn{1}{c|}{\multirow{3}{*}{Multi-source DA}} &
  \multicolumn{1}{c|}{SImpAl} &
  46.8 &
  56.7 &
  65.1 &
  66.6 &
  \multicolumn{1}{c|}{58.8} &
  49.3 &
  62.1 &
  71.7 &
  73.0 &
  64.1 \\
\multicolumn{1}{c|}{} &
  \multicolumn{1}{c|}{} &
  \multicolumn{1}{c|}{MFSAN} &
  39.9 &
  46.6 &
  58.9 &
  55.6 &
  \multicolumn{1}{c|}{50.3} &
  44.5 &
  53.7 &
  65.4 &
  64.2 &
  57.0 \\
\multicolumn{1}{c|}{} &
  \multicolumn{1}{c|}{} &
  \multicolumn{1}{c|}{PMDA} &
  50.8 &
  56.8 &
  64.2 &
  66.8 &
  \multicolumn{1}{c|}{59.7} &
  54.4 &
  65.8 &
  70.4 &
  71.8 &
  65.6 \\ \cline{2-13} 
\multicolumn{1}{c|}{} &
  \multicolumn{1}{c|}{MFDA} &
  \multicolumn{1}{c|}{MSFAN} &
  55.6 &
  68.4 &
  75.6 &
  76.6 &
  \multicolumn{1}{c|}{69.1} &
  56.3 &
  68.7 &
  79.3 &
  79.1 &
  70.9 \\ \hline
\multicolumn{1}{c|}{\multirow{11}{*}{\begin{tabular}[c]{@{}c@{}}ViT-B/16\\ CLIP (*)\end{tabular}}} &
  \multicolumn{1}{c|}{Zero-shot} &
  \multicolumn{1}{c|}{CLIP} &
  67.7 &
  82.7 &
  89.2 &
  89.6 &
  \multicolumn{1}{c|}{82.3} &
  67.7 &
  82.7 &
  89.2 &
  89.6 &
  82.3 \\ \cline{2-13} 
\multicolumn{1}{c|}{} &
  \multicolumn{1}{c|}{\multirow{4}{*}{Single-best}} &
  \multicolumn{1}{c|}{CoOP} &
  70.0±0.3 &
  82.4±0.6 &
  90.9±0.2 &
  90.3±0.3 &
  \multicolumn{1}{c|}{83.4±0.2} &
  70.3±0.6 &
  82.5±0.4 &
  90.6±0.4 &
  90.4±0.8 &
  83.5±0.5 \\
\multicolumn{1}{c|}{} &
  \multicolumn{1}{c|}{} &
  \multicolumn{1}{c|}{VPT} &
  69.6±0.4 &
  83.0±0.5 &
  90.2±0.2 &
  90.2±0.1 &
  \multicolumn{1}{c|}{83.3±0.2} &
  69.7±0.4 &
  83.9±0.6 &
  90.5±0.4 &
  90.3±0.4 &
  83.6±0.3 \\
\multicolumn{1}{c|}{} &
  \multicolumn{1}{c|}{} &
  \multicolumn{1}{c|}{MaPLe} &
  71.1±0.7 &
  83.3±0.5 &
  91.2±0.2 &
  90.5±0.6 &
  \multicolumn{1}{c|}{84.0±0.4} &
  71.0±0.6 &
  83.1±0.5 &
  91.8±0.4 &
  90.7±0.2 &
  84.1±0.3 \\
  \multicolumn{1}{c|}{} &
  \multicolumn{1}{c|}{} &
  \multicolumn{1}{c|}{MaPLe$_t$} &
  71.7±0.4 &
  84.3±0.1 &
  \textbf{91.7}±0.2 &
  90.7±0.2 &
  \multicolumn{1}{c|}{84.6±0.1} &
  72.6±0.4 &
  84.5±0.2 &
  91.9±0.4 &
  91.1±0.3 &
  85.0±0.1 \\ \cline{2-13} 
\multicolumn{1}{c|}{} &
  \multicolumn{1}{c|}{\multirow{4}{*}{Source-combined}} &
  \multicolumn{1}{c|}{CoOP} &
  71.6±0.6 &
  82.8±0.5 &
  90.7±0.5 &
  89.7±0.3 &
  \multicolumn{1}{c|}{83.7±0.3} &
  71.0±0.4 &
  83.3±0.4 &
  91.7±0.5 &
  90.7±0.2 &
  84.2±0.1 \\
\multicolumn{1}{c|}{} &
  \multicolumn{1}{c|}{} &
  \multicolumn{1}{c|}{VPT} &
  71.0±0.5 &
  83.8±0.1 &
  90.3±0.1 &
  90.3±0.1 &
  \multicolumn{1}{c|}{83.8±0.1} &
  70.9±0.9 &
  83.8±0.4 &
  90.3±0.2 &
  90.4±0.1 &
  83.8±0.2 \\
\multicolumn{1}{c|}{} &
  \multicolumn{1}{c|}{} &
  \multicolumn{1}{c|}{MaPLe} &
  71.3±0.5 &
  83.1±0.5 &
  91.2±0.4 &
  90.4±0.4 &
  \multicolumn{1}{c|}{84.0±0.1} &
  72.2±0.5 &
  83.2±0.4 &
  \textbf{92.0}±0.2 &
  90.6±0.2 &
  84.5±0.1 \\ 
  \multicolumn{1}{c|}{} &
  \multicolumn{1}{c|}{} &
  \multicolumn{1}{c|}{MaPLe$_t$} &
  71.5±0.6 &
  84.2±0.2 &
  91.4±0.4 &
  \textbf{91.1}±0.3 &
  \multicolumn{1}{c|}{84.5±0.2} &
  71.7±0.2 &
  84.5±0.4 &
  91.9±0.1 &
  91.0±0.3 &
  84.8±0.1 \\ \cline{2-13}
\multicolumn{1}{c|}{} &
  \multicolumn{1}{c|}{\multirow{2}{*}{Single-best DA}} &
  \multicolumn{1}{c|}{DAPL} &
  70.0±0.1 &
  83.5±0.7 &
  91.0±0.5 &
  90.5±0.2 &
  \multicolumn{1}{c|}{83.7±0.4} &
  70.2±0.3 &
  84.3±0.3 &
  90.9±0.4 &
  90.6±0.2 &
  84.0±0.1 \\
\multicolumn{1}{c|}{} &
   \multicolumn{1}{c|}{} &
  \multicolumn{1}{c|}{DAPL-CSC} &
   68.8±1.0 &
  83.2±0.4 &
  90.3±0.7 &
  90.1±0.3 &
  \multicolumn{1}{c|}{83.1±0.4} &
  69.5±1.2 &
  83.7±0.3 &
  90.6±0.8 &
  90.4±0.2 &
  83.5±0.4 \\ \cline{2-13}
\multicolumn{1}{c|}{} &
  \multicolumn{1}{c|}{\multirow{2}{*}{Source-combined DA}} &
  \multicolumn{1}{c|}{DAPL} &
  63.0±4.5 &
  74.2±4.9 &
  83.2±5.7 &
  84.3±4.4 &
  \multicolumn{1}{c|}{76.2±4.8} &
  62.3±2.2 &
  74.8±3.0 &
  84.3±4.2 &
  83.2±3.9 &
  76.1±1.9 \\
  \multicolumn{1}{c|}{} &
   \multicolumn{1}{c|}{} &
  \multicolumn{1}{c|}{DAPL-CSC} &
   39.2±3.2 &
  40.4±3.0 &
  51.9±3.4 &
  44.7±2.4 &
  \multicolumn{1}{c|}{44.0±3.0} &
  37.1±2.8 &
  37.6±1.7 &
  50.7±4.8 &
  45.6±1.6 &
  42.7±1.7 \\ \cline{2-13} 
\multicolumn{1}{c|}{} &
  \multicolumn{1}{c|}{\multirow{2}{*}{Multi-source DA}} &
  \multicolumn{1}{c|}{MPA} &
   63.0±0.5 &
  76.9±1.3 &
  83.5±1.1 &
  81.6±0.4 &
  \multicolumn{1}{c|}{76.2±0.2} &
  63.5±0.7 &
  77.3±0.9 &
  83.4±0.3 &
  81.2±0.5 &
  76.3±0.3 \\
\multicolumn{1}{c|}{} &
  \multicolumn{1}{c|}{} &
  \multicolumn{1}{c|}{MPA-CSC} &
  63.2±0.7 &
  77.2±0.5 &
  83.9±0.4 &
  81.7±0.0 &
  \multicolumn{1}{c|}{76.5±0.1} &
  62.4±1.0 &
  76.8±0.5 &
  82.7±1.4 &
  80.4±1.2 &
  75.6±0.7 \\ \cline{2-13} 
\multicolumn{1}{c|}{} &
  \multicolumn{1}{c|}{MFDA} &
  \multicolumn{1}{c|}{VAMP} &
  \textbf{73.7}±0.6 &
  \textbf{85.7}±0.3 &
  91.4±0.4 &
  90.9±0.2 &
  \multicolumn{1}{c|}{\textbf{85.4}±0.2} &
  \textbf{73.5}±0.2 &
  \textbf{85.8}±0.4 &
  91.4±0.1 &
  \textbf{91.4}±0.2 &
  \textbf{85.5}±0.1 \\ \hline
\end{tabular}
}
\caption{Accuracy evaluation (\%) and Standard deviation (\%) on target domain of OfficeHome dataset. (*) denotes that the method is based on our implementation.}
\label{tab:tab6}
\end{table*}
\begin{table*}[!t]
\centering
\resizebox{\linewidth}{!}{
\begin{tabular}{ccccccccccccc}
\hline
\multicolumn{13}{c}{DomainNet} \\ \hline
\multicolumn{1}{c|}{\multirow{3}{*}{Backbone}} &
  \multicolumn{2}{c|}{\multirow{3}{*}{Method}} &
  \multicolumn{5}{c|}{1 shot} &
  \multicolumn{5}{c}{3 shot} \\ \cline{4-13} 
\multicolumn{1}{c|}{} &
  \multicolumn{2}{c|}{} &
  \multirow{2}{*}{\begin{tabular}[c]{@{}c@{}}PRS\\ →C\end{tabular}} &
  \multirow{2}{*}{\begin{tabular}[c]{@{}c@{}}CRS\\ →P\end{tabular}} &
  \multirow{2}{*}{\begin{tabular}[c]{@{}c@{}}CPS\\ →R\end{tabular}} &
  \multirow{2}{*}{\begin{tabular}[c]{@{}c@{}}CPR\\ →S\end{tabular}} &
  \multicolumn{1}{c|}{\multirow{2}{*}{Avg}} &
  \multirow{2}{*}{\begin{tabular}[c]{@{}c@{}}PRS\\ →C\end{tabular}} &
  \multirow{2}{*}{\begin{tabular}[c]{@{}c@{}}CRS\\ →P\end{tabular}} &
  \multirow{2}{*}{\begin{tabular}[c]{@{}c@{}}CPS\\ →R\end{tabular}} &
  \multirow{2}{*}{\begin{tabular}[c]{@{}c@{}}CPR\\ →S\end{tabular}} &
  \multirow{2}{*}{Avg} \\
\multicolumn{1}{c|}{} &
  \multicolumn{2}{c|}{} &
   &
   &
   &
   &
  \multicolumn{1}{c|}{} &
   &
   &
   &
   &
   \\ \hline
\multicolumn{1}{c|}{\multirow{16}{*}{ResNet-101}} &
  \multicolumn{2}{c|}{Single-best} &
  18.4 &
  30.6 &
  28.9 &
  16.7 &
  \multicolumn{1}{c|}{23.7} &
  30.2 &
  44.2 &
  49.8 &
  24.2 &
  34.4 \\ \cline{2-13} 
\multicolumn{1}{c|}{} &
  \multicolumn{2}{c|}{Source-combined} &
  30.8 &
  49.4 &
  43.3 &
  36.9 &
  \multicolumn{1}{c|}{40.1} &
  45.3 &
  57.4 &
  64.7 &
  42.6 &
  50.0 \\ \cline{2-13} 
\multicolumn{1}{c|}{} &
  \multicolumn{1}{c|}{\multirow{5}{*}{Single-best DA}} &
  \multicolumn{1}{c|}{CDAN} &
  16.0 &
  25.7 &
  19.5 &
  12.9 &
  \multicolumn{1}{c|}{18.5} &
  30.0 &
  40.1 &
  40.8 &
  17.1 &
  29.3 \\
\multicolumn{1}{c|}{} &
  \multicolumn{1}{c|}{} &
  \multicolumn{1}{c|}{MME} &
  16.0 &
  29.2 &
  26.0 &
  13.4 &
  \multicolumn{1}{c|}{21.2} &
  25.1 &
  46.5 &
  50.0 &
  20.1 &
  32.6 \\
\multicolumn{1}{c|}{} &
  \multicolumn{1}{c|}{} &
  \multicolumn{1}{c|}{MDDIA} &
  18.0 &
  30.6 &
  27.4 &
  15.9 &
  \multicolumn{1}{c|}{23} &
  41.4 &
  50.7 &
  52.9 &
  23.1 &
  38.2 \\
\multicolumn{1}{c|}{} &
  \multicolumn{1}{c|}{} &
  \multicolumn{1}{c|}{CDS} &
  16.7 &
  24.4 &
  15.9 &
  13.4 &
  \multicolumn{1}{c|}{17.6} &
  35.0 &
  43.8 &
  36.8 &
  31.1 &
  32.9 \\
\multicolumn{1}{c|}{} &
  \multicolumn{1}{c|}{} &
  \multicolumn{1}{c|}{PCS} &
  39.0 &
  51.7 &
  38.8 &
  39.8 &
  \multicolumn{1}{c|}{42.3} &
  45.2 &
  59.1 &
  66.6 &
  41.9 &
  51.0 \\ \cline{2-13} 
\multicolumn{1}{c|}{} &
  \multicolumn{1}{c|}{\multirow{5}{*}{Source-combined DA}} &
  \multicolumn{1}{c|}{CDAN} &
  25.7 &
  33.0 &
  40.0 &
  26.4 &
  \multicolumn{1}{c|}{31.3} &
  47.8 &
  54.1 &
  65.6 &
  49.1 &
  49.6 \\
\multicolumn{1}{c|}{} &
  \multicolumn{1}{c|}{} &
  \multicolumn{1}{c|}{MME} &
  20.0 &
  45.3 &
  52.5 &
  13.0 &
  \multicolumn{1}{c|}{32.7} &
  44.2 &
  62.7 &
  73.9 &
  51.8 &
  53.1 \\
\multicolumn{1}{c|}{} &
  \multicolumn{1}{c|}{} &
  \multicolumn{1}{c|}{MDDIA} &
  44.0 &
  46.4 &
  49.6 &
  37.1 &
  \multicolumn{1}{c|}{44.3} &
  56.3 &
  59.3 &
  70.3 &
  51.3 &
  56.3 \\
\multicolumn{1}{c|}{} &
  \multicolumn{1}{c|}{} &
  \multicolumn{1}{c|}{CDS} &
  42.2 &
  53.3 &
  55.4 &
  38.5 &
  \multicolumn{1}{c|}{47.4} &
  50.2 &
  61.5 &
  71.8 &
  47.3 &
  55.6 \\
\multicolumn{1}{c|}{} &
  \multicolumn{1}{c|}{} &
  \multicolumn{1}{c|}{PCS} &
  36.2 &
  53.0 &
  56.4 &
  32.8 &
  \multicolumn{1}{c|}{44.6} &
  45.6 &
  61.2 &
  74.3 &
  41.3 &
  53.4 \\ \cline{2-13} 
\multicolumn{1}{c|}{} &
  \multicolumn{1}{c|}{\multirow{3}{*}{Multi-source DA}} &
  \multicolumn{1}{c|}{SImpAl} &
  48.0 &
  40.3 &
  45.7 &
  35.3 &
  \multicolumn{1}{c|}{42.3} &
  51.5 &
  47.4 &
  68.8 &
  45.3 &
  51.1 \\
\multicolumn{1}{c|}{} &
  \multicolumn{1}{c|}{} &
  \multicolumn{1}{c|}{MFSAN} &
  41.6 &
  33.5 &
  38.8 &
  29.6 &
  \multicolumn{1}{c|}{35.9} &
  43.5 &
  42.3 &
  63.2 &
  41.1 &
  45.2 \\
\multicolumn{1}{c|}{} &
  \multicolumn{1}{c|}{} &
  \multicolumn{1}{c|}{PMDA} &
  49.3 &
  42.2 &
  45.0 &
  34.8 &
  \multicolumn{1}{c|}{42.8} &
  52.2 &
  52.5 &
  71.3 &
  47.6 &
  53.3 \\ \cline{2-13} 
\multicolumn{1}{c|}{} &
  \multicolumn{1}{c|}{MFDA} &
  \multicolumn{1}{c|}{MSFAN} &
  57.3 &
  68.7 &
  64.8 &
  45.2 &
  \multicolumn{1}{c|}{59.0} &
  57.8 &
  65.5 &
  75.8 &
  53.6 &
  62.3 \\ \hline
\multicolumn{1}{c|}{\multirow{11}{*}{\begin{tabular}[c]{@{}c@{}}ViT-B/16 \\ CLIP (*)\end{tabular}}} &
  \multicolumn{1}{c|}{Zero-shot} &
  \multicolumn{1}{c|}{CLIP} &
  82.7 &
  82.6 &
  91.8 &
  79.6 &
  \multicolumn{1}{c|}{84.2} &
  82.7 &
  82.6 &
  91.8 &
  79.6 &
  84.2 \\ \cline{2-13} 
\multicolumn{1}{c|}{} &
  \multicolumn{1}{c|}{\multirow{4}{*}{Single-best}} &
  \multicolumn{1}{c|}{CoOP} &
  82.3±0.6 &
  81.9±0.5 &
  90.9±0.4 &
  79.1±0.3 &
  \multicolumn{1}{c|}{83.5±0.4} &
  83.5±0.2 &
  82.8±0.3 &
  91.1±0.1 &
  80.0±0.6 &
  84.3±0.3 \\
\multicolumn{1}{c|}{} &
  \multicolumn{1}{c|}{} &
  \multicolumn{1}{c|}{VPT} &
  82.4±0.2 &
  82.2±0.3 &
  91.7±0.0 &
  80.0±0.2 &
  \multicolumn{1}{c|}{84.0±0.1} &
  83.4±0.1 &
  83.1±0.1 &
  91.9±0.1 &
  80.3±0.0 &
  84.7±0.0 \\
\multicolumn{1}{c|}{} &
  \multicolumn{1}{c|}{} &
  \multicolumn{1}{c|}{MaPLe} &
  83.3±0.6 &
  82.9±0.0 &
  91.8±0.3 &
  79.6±0.2 &
  \multicolumn{1}{c|}{84.4±0.1} &
  84.6±0.3 &
  83.3±0.2 &
  91.6±0.2 &
  80.1±0.5 &
  84.9±0.1 \\ 
  \multicolumn{1}{c|}{} &
  \multicolumn{1}{c|}{} &
  \multicolumn{1}{c|}{MaPLe$_t$} &
  83.5±0.4 &
  82.8±0.3 &
  91.9±0.1 &
  80.4±0.5 &
  \multicolumn{1}{c|}{84.6±0.1} &
  84.9±0.2 &
  83.7±0.2 &
  92.1±0.1 &
  81.1±0.4 &
  85.4±0.1 \\ \cline{2-13} 
\multicolumn{1}{c|}{} &
  \multicolumn{1}{c|}{\multirow{4}{*}{Source-combined}} &
  \multicolumn{1}{c|}{CoOP} &
  83.0±0.5 &
  82.4±0.5 &
  91.3±0.1 &
  79.5±0.3 &
  \multicolumn{1}{c|}{84.0±0.1} &
  84.0±0.3 &
  84.1±0.1 &
  91.4±0.4 &
  80.8±0.2 &
  85.1±0.1 \\
\multicolumn{1}{c|}{} &
  \multicolumn{1}{c|}{} &
  \multicolumn{1}{c|}{VPT} &
  82.7±0.4 &
  82.3±0.6 &
  91.9±0.1 &
  79.6±0.5 &
  \multicolumn{1}{c|}{84.1±0.3} &
  83.8±0.1 &
  83.5±0.3 &
  92.0±0.1 &
  80.7±0.3 &
  85.0±0.1 \\
\multicolumn{1}{c|}{} &
  \multicolumn{1}{c|}{} &
  \multicolumn{1}{c|}{MaPLe} &
  83.8±0.4 &
  83.2±0.4 &
  91.9±0.3 &
  79.9±0.7 &
  \multicolumn{1}{c|}{84.7±0.1} &
  84.8±0.2 &
  84.0±0.2 &
  91.7±0.3 &
  80.4±0.8 &
  85.2±0.3 \\ 
  \multicolumn{1}{c|}{} &
  \multicolumn{1}{c|}{} &
  \multicolumn{1}{c|}{MaPLe$_t$} &
  \textbf{84.3}±0.1 &
  83.8±0.4 &
  92.3±0.0 &
  80.7±0.3 &
  \multicolumn{1}{c|}{85.3±0.1} &
  84.7±0.2 &
  84.4±0.1 &
  \textbf{92.7}±0.1 &
  81.4±0.1 &
  85.8±0.1 \\ \cline{2-13}
\multicolumn{1}{c|}{} &
  \multicolumn{1}{c|}{\multirow{2}{*}{Single-best DA}} &
  \multicolumn{1}{c|}{DAPL} &
  83.4±0.8 &
  83.7±0.3 &
  91.9±0.4 &
  \textbf{80.8}±0.1 &
  \multicolumn{1}{c|}{85.0±0.4} &
  83.6±0.4 &
  84.3±0.2 &
  92.1±0.5 &
  81.2±0.2 &
  85.3±0.2 \\
\multicolumn{1}{c|}{} &
   \multicolumn{1}{c|}{} &
  \multicolumn{1}{c|}{DAPL-CSC} &
   83.5±0.1 &
  83.2±0.2 &
  91.8±0.2 &
  80.2±0.4 &
  \multicolumn{1}{c|}{84.7±0.1} &
  83.9±0.4 &
  83.9±0.2 &
  92.1±0.2 &
  80.8±0.2 &
  85.2±0.1 \\ \cline{2-13}
\multicolumn{1}{c|}{} &
  \multicolumn{1}{c|}{\multirow{2}{*}{Source-combined DA}} &
  \multicolumn{1}{c|}{DAPL} &
  80.2±0.7 &
  75.6±3.4 &
  87.3±4.4 &
  77.3±2.4 &
  \multicolumn{1}{c|}{80.1±1.8} &
  80.0±3.0 &
  79.3±1.6 &
  89.0±1.6 &
  78.6±1.2 &
  80.1±1.7 \\
\multicolumn{1}{c|}{} &
   \multicolumn{1}{c|}{} &
  \multicolumn{1}{c|}{DAPL-CSC} &
  37.6±3.9 &
  29.6±2.7 &
  33.2±3.0 &
  28.7±1.6 &
  \multicolumn{1}{c|}{32.3±2.7} &
  34.5±3.8 &
  27.4±2.4 &
  30.9±1.0 &
  27.0±2.8 &
  29.9±2.1 \\ \cline{2-13}
\multicolumn{1}{c|}{} &
  \multicolumn{1}{c|}{\multirow{2}{*}{Multi-source DA}} &
  \multicolumn{1}{c|}{MPA} &
  82.5±1.1 &
  82.5±0.2 &
  91.4±0.1 &
  80.2±0.1 &
  \multicolumn{1}{c|}{84.2±0.2} &
  83.5±0.2 &
  82.6±0.2 &
  91.8±0.0 &
  80.4±0.3 &
  84.4±0.1 \\
\multicolumn{1}{c|}{} &
 \multicolumn{1}{c|}{} &
  \multicolumn{1}{c|}{MPA-CSC} &
  83.2±0.2 &
  83.0±0.3 &
  91.1±0.1 &
  80.3±0.1 &
  \multicolumn{1}{c|}{84.4±0.1} &
  83.3±0.4 &
  82.8±0.2 &
  91.1±0.2 &
  80.5±0.4 &
  84.4±0.2 \\ \cline{2-13} 
\multicolumn{1}{c|}{} &
  \multicolumn{1}{c|}{MFDA} &
  \multicolumn{1}{c|}{VAMP} &
  \textbf{84.3}±0.1 &
  \textbf{84.3}±0.1 &
  \textbf{92.6}±0.1 &
  \textbf{80.8}±0.2 &
  \multicolumn{1}{c|}{\textbf{85.5}±0.0} &
  \textbf{85.1}±0.1 &
  \textbf{84.9}±0.1 &
  92.5±0.0 &
  \textbf{81.9}±0.0 &
  \textbf{86.1}±0.0 \\ \hline
\end{tabular}
}
\caption{Accuracy evaluation (\%) and Standard deviation (\%) on target domain of DomainNet dataset.}
\label{tab:tab7}
\end{table*}
\begin{enumerate}[label=(\arabic*)]
    \item \textbf{Zero-shot.} It denotes that CLIP zero-shot inference on the target domain data is implemented.
    \item The direct full-parameter finetuning method and the efficiently prompt tuning method (VPT (Only vision branch tuned), CoOp \cite{Zhou2022b}, MaPLe \cite{Khattak2023}) can be conducted in the following two types:
    \begin{itemize}
        \item \textbf{Single-best.} It denotes the model is directly trained on a small number of labeled data of each source domain, respectively, and then evaluated on the target domain data. The best result of training among source domains is reported.
        \item \textbf{Source-combined.}  It denotes that the model is directly trained on all the annotated data that combines all source domains.
    \end{itemize}
    \item Some baseline methods were originally proposed for single-source DA. The conventional methods include CDAN \cite{Long2018}, MMDIA \cite{Jiang2020}, MME \cite{Saito2019}, CDS \cite{Kim2020} and PCS \cite{Yue2021b}, where CDS and PCS are customized for single-source few-shot DA (FUDA) that only has small annotated source data to provide. For the prompt-based single-source DA methods, DAPL \cite{Ge2022} utilizes disentangling prompt tuning. The above methods are re-implemented in MFDA categorizing the two types: 
    \begin{itemize}
        \item \textbf{Single-best DA.} Similar to Single-best, it is trained using a small number of labeled data of each source domain by the single-source DA method and evaluated on the target domain data. The best result of training among source domains is reported.
        \item \textbf{Source-combined DA.} Similar to Source-combined, it is trained on all the annotated data and treats all the source domains as one source domain. Then the single-source DA method is adopted to compare.
    \end{itemize}
    \item \textbf{Multi-source DA.} These methods are designed for the multi-source domain adaptation assuming that fully-annotated labels from multiple source domains are available. It includes the full-parameter tuning methods: SImpAl \cite{Venkat2020}), MFSAN \cite{Zhu2019}, PMDA \cite{Zhou2022} and prompt-based method: MPA \cite{Chen2023}. All of them are retrained under the few annotated source samples in the MFDA setting.
\end{enumerate}
The comparative results are shown in Table. \ref{tab:tab6} and Table. \ref{tab:tab7}. MaPLe$_t$ means that we further add the qualified pseudo label of target samples for training of MaPLe, such as Eq.(\ref{eq:eq8}). The comparison results with ours show that it is not enough to use the pseudo-label training of target samples as a supplement. Our VAMP also considers the alignment of the domain distribution, which adds cross-domain knowledge. DAPL-CSC and MPA-CSC denote that the domain-agnostic context follows a class-specific context. That means each class would be initialized with different tokens, which requires more tunable parameters. As indicated, the performance of those with CSC is relatively low compared with those without CSC in the Single-best DA, and even their performance has a significant degradation in the Source-combined DA. Additionally, the proposed VAMP framework achieves competitive results in the MFDA setting, further demonstrating our VAMP's effectiveness. It can be observed that prompt tuning methods can outperform conventional methods. However, similar to MaPLe, our VAMP involves prompt tuning of the visual encoder based on Transformer \cite{Dosovitskiy2021}, which does not apply to ResNet. There may be some degree of unfairness compared to conventional methods, because of the differences in the capabilities and parameter scales of the different vision encoders.

\end{document}